\documentclass[11pt,letterpaper]{article}

\usepackage[letterpaper,top=0.72in,bottom=0.72in,left=0.88in,right=0.88in]{geometry}
\usepackage{fourier}
\usepackage[T1]{fontenc}
\usepackage[utf8]{inputenc}
\usepackage{graphicx}
\usepackage{booktabs}
\usepackage{multirow}
\usepackage{tabularx}
\usepackage{array}
\usepackage{enumitem}
\usepackage{caption}
\usepackage{subcaption}
\usepackage{float}
\usepackage{placeins}
\usepackage{amsmath,amssymb}
\usepackage{xcolor}
\usepackage{tikz}
\usetikzlibrary{arrows.meta,positioning,calc,fit,shapes.geometric}
\usepackage[most]{tcolorbox}
\usepackage{listings}
\usepackage[backend=biber,style=numeric,sorting=none,sortcites=true,natbib=true]{biblatex}
\AtBeginBibliography{\small}
\usepackage[colorlinks=true,allcolors=ExaonePurpleDark]{hyperref}
\usepackage[capitalise,noabbrev]{cleveref}
\usepackage{titlesec}

\usepackage[T1]{fontenc}
\usepackage[outline]{contour}
\usepackage{adjustbox}

\usepackage{pdfrender}

\newcommand{\ReportTitle}{EXAONE Tabular 1.0 : Technical Report}
\newcommand{\ReportSubtitle}{A Synthetic-Prior Foundation Model for Tabular Classification and Regression}

\newcommand{\EqualContribMark}{%
  \begingroup
  \renewcommand{\thefootnote}{\fnsymbol{footnote}}%
  \footnotemark[1]%
  \endgroup
}

\newcommand{\ReportAuthors}{
  Moonjung Eo\EqualContribMark,
  Min-Kook Suh\EqualContribMark,
  Hye-Seung Cho\EqualContribMark,
  Jiwon Kim\EqualContribMark,
  Seoyoon Kim\EqualContribMark,
  Sangjun Nam\EqualContribMark,
  Soonyoung Lee
}

\newcommand{\ReportAffiliation}{LG AI Research}
\newcommand{\ReportDate}{August 2026}
\newcommand{\RepositoryURL}{https://github.com/LGAI-Research/EXAONE-Tabular}
\newcommand{\ModelCardURL}{https://huggingface.co/LG-AI-Research/EXAONE-Tabular}

\definecolor{ExaonePurple}{HTML}{8F3FE6}
\definecolor{ExaonePurpleDark}{HTML}{5A1FA8}
\definecolor{ExaonePurpleLight}{HTML}{EADAFB}
\definecolor{ExaoneBlue}{HTML}{5D6FEF}
\definecolor{OneGray}{HTML}{F6F6F7}
\definecolor{OneGrayMid}{HTML}{D8D8DD}
\definecolor{OneGrayDark}{HTML}{55555D}
\definecolor{TodoBack}{HTML}{FFF8E8}
\definecolor{TodoFrame}{HTML}{D7A12A}

\setlist[itemize]{leftmargin=1.35em,itemsep=0.15em,topsep=0.25em}
\setlist[enumerate]{leftmargin=1.55em,itemsep=0.2em,topsep=0.25em}

\newcolumntype{Y}{>{\raggedright\arraybackslash}X}
\newcolumntype{R}{>{\raggedleft\arraybackslash}X}
\newcolumntype{C}{>{\centering\arraybackslash}X}

\newcommand{\EXAONETabular}{{\fontfamily{qcs}\selectfont EXAONE\nobreakspace{}Tabular}}   
\newcommand{\code}[1]{\texttt{#1}}

\newtcolorbox{todobox}[1][]{
  enhanced,
  breakable,
  colback=TodoBack,
  colframe=TodoFrame,
  boxrule=0.6pt,
  arc=1.5mm,
  left=1.7mm,right=1.7mm,top=1.2mm,bottom=1.2mm,
  fonttitle=\bfseries,
  title={TODO --- final measurement},
  #1
}

\newtcolorbox{releasebox}{
  enhanced,
  colback=OneGray,
  colframe=OneGrayMid,
  boxrule=0.4pt,
  arc=4mm,
  left=5mm,right=5mm,top=5mm,bottom=4mm
}

\lstdefinestyle{exaonetabular}{
  basicstyle=\ttfamily\small,
  backgroundcolor=\color{OneGray},
  frame=single,
  rulecolor=\color{OneGrayMid},
  framerule=0.4pt,
  breaklines=true,
  columns=fullflexible,
  keepspaces=true,
  showstringspaces=false,
  xleftmargin=0.6em,
  xrightmargin=0.6em,
  aboveskip=0.5em,
  belowskip=0.5em
}

\titleformat{\section}{\large\bfseries}{\thesection}{0.75em}{}
\titleformat{\subsection}{\normalsize\bfseries}{\thesubsection}{0.65em}{}
\titleformat{\subsubsection}{\normalsize\bfseries}{\thesubsubsection}{0.6em}{}
\titlespacing*{\section}{0pt}{0.95em}{0.35em}
\titlespacing*{\subsection}{0pt}{0.65em}{0.22em}
\titlespacing*{\subsubsection}{0pt}{0.5em}{0.18em}

\begin{document}

\thispagestyle{empty}
\begin{releasebox}
  \noindent
  \begin{minipage}[c]{0.13\linewidth}
    \centering
    \includegraphics[height=0.8cm,keepaspectratio]{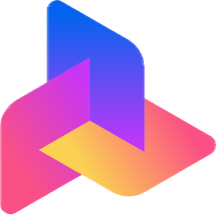}
  \end{minipage}%
  \begin{minipage}[c]{0.82\linewidth}
    {\fontsize{19}{22}\selectfont\bfseries
      \ReportTitle\par
    }
  \end{minipage}
  \par
  \vspace{0.30em}
  {\fontsize{13.2}{15.5}\selectfont\bfseries
    \ReportSubtitle\par
  }
  \vspace{0.45em}
  {\large\bfseries
    \ReportAuthors\par
  }
  \vspace{0.08em}
  {\normalsize
    \ReportAffiliation\par
  }
  \vspace{0.85em}
  
  \noindent
\EXAONETabular{} is a compact tabular foundation model family for classification and regression via in-context learning, producing predictions without dataset-specific gradient updates. Pretrained exclusively on a synthetic structural-causal-model (SCM) prior, its central contribution is an architecture-centered redesign of tabular in-context learning. Rather than compressing features into a fixed row embedding before a separate row-level learner, \EXAONETabular{} interleaves feature-axis attention within each item with support-conditioned item-axis attention within each feature at every Transformer layer, mediated by item-summary and feature-summary tokens. Across four public benchmarks, \EXAONETabular{} combines strong predictive performance with high efficiency. On TabArena, its 20.81M-parameter classification model ranks first overall, surpassing tuned ensembles and 4-hour AutoML pipelines, while regression reaches the performance regime of the 1.64B-parameter TabFM at roughly 1/11 the inference cost. On BCCO and TALENT, \EXAONETabular{} ranks second in classification and first in regression. On ScoringBench, it achieves the best mean rank for both point-estimation and predictive-distribution quality, leading the $R^2$, RMSE, and CRPS evaluations. Together, these results establish \EXAONETabular{} as a state-of-the-art compact tabular foundation model family, combining strong predictive performance across classification, point regression, and probabilistic regression with an efficient model design.


  \vspace{0.75em}
  \begin{tabularx}{\linewidth}{@{}>{\bfseries}p{0.92in}Y@{}}
    Model: & \EXAONETabular{} \\
    Date: & \ReportDate \\
    GitHub: & \RepositoryURL \\
    Hugging Face: & \ModelCardURL \\
  \end{tabularx}
\end{releasebox}
\begingroup
\renewcommand{\thefootnote}{\fnsymbol{footnote}}
\footnotetext[1]{These authors contributed equally.}
\endgroup



\begin{figure}[h]
  \centering

  \includegraphics[
    width=.9\textwidth,
    height=0.28\textheight,
    keepaspectratio
  ]{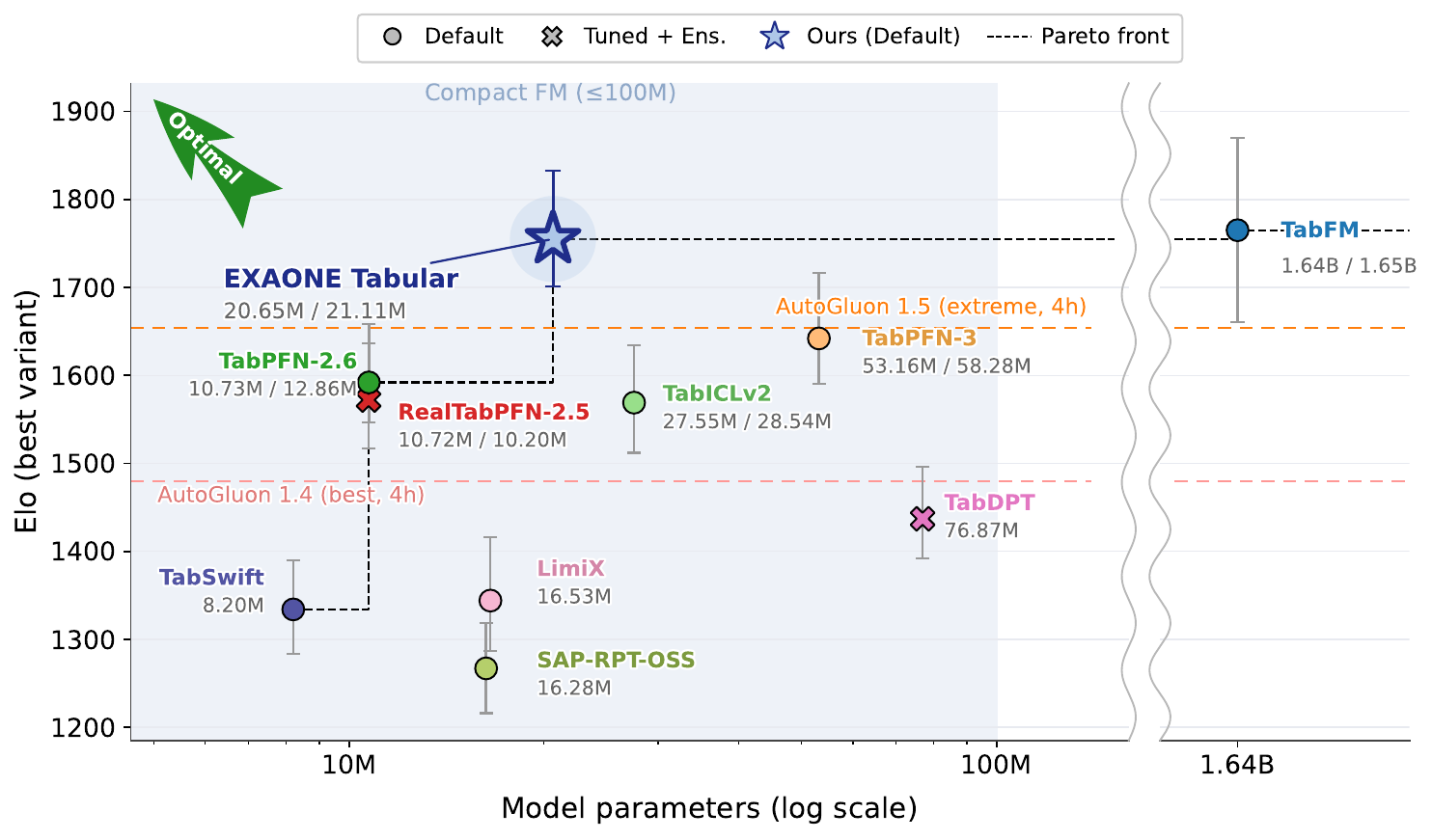}

\caption{Performance--size Pareto front on the full TabArena benchmark (classification and regression). \EXAONETabular{} lies on the Pareto front in its default setting and is the strongest compact tabular foundation model ($\leq$100M parameters), outperforming every model shown except TabFM --- with which it is statistically tied at $\sim$1.3\% of the parameters --- by 113 to 488 Elo. Sizes count the weights stored in the evaluated checkpoints.}
  \label{fig:size-performance-pareto}
\end{figure}

\clearpage
\setcounter{page}{2}

\section{Introduction}

Tabular prediction supports operational decisions in clinical risk estimation~\cite{pham2026retrieval}, credit scoring~\cite{clements2020sequential}, predictive maintenance~\cite{hector2024predictive}, manufacturing quality control~\cite{md2022review}, scientific measurement~\cite{attari2025decoding}, and many other real-world domains~\cite{borisov2022deep}. For many years, gradient-boosted decision trees, including XGBoost, LightGBM, and CatBoost, have been the strongest general-purpose default for tabular data~\citep{chen2016xgboost,ke2017lightgbm,prokhorenkova2018catboost,grinsztajn2022why}. However, applying these models involves separate training for each dataset. Such dataset-specific training requires labeled data and computational resources for every new task, making deployment costly and time-consuming when rapid adaptation across many datasets is required.

Tabular foundation models challenge this convention by pretraining a reusable prediction algorithm across many tasks and applying it to a new dataset through in-context learning~\citep{muller2022transformers,hollmann2023tabpfn,hollmann2025accurate,qu2025tabicl,grinsztajn2026tabpfn3,qu2026tabiclv2}. Instead of optimizing a new set of parameters for each dataset, a pretrained model receives a labeled support set together with an unlabeled query set and directly predicts the corresponding query targets. Formally, given a labeled support set
\begin{equation}
D_{\mathrm{support}}=\{(x_i,y_i)\}_{i=1}^{N},
\end{equation}
and an unlabeled query set
\begin{equation}
X_{\mathrm{query}}=\{x_j\}_{j=N+1}^{N+M},
\end{equation}
the model predicts the query targets \(Y_{\mathrm{query}}=\{y_j\}_{j=N+1}^{N+M}\) through
\begin{equation}
p\!\left(Y_{\mathrm{query}}\mid X_{\mathrm{query}},D_{\mathrm{support}}\right),
\end{equation}
without updating its pretrained parameters. By meta-training over a broad family of data-generation mechanisms, prior-fitted models can therefore produce predictions on previously unseen datasets through forward inference rather than dataset-specific gradient-based optimization~\citep{muller2022transformers}.

\EXAONETabular{} is LG AI Research's first tabular foundation model family, comprising separately trained classification and regression models that share the same core architecture and are pretrained from scratch on synthetic data generated from a structural causal model (SCM) prior. At the core of both models is the \emph{Cross-Axis Summary Transformer} (CAST), which maintains a separate representation for each table cell throughout the network. Each CAST layer first applies feature-axis attention within each item and then performs support-conditioned item-axis attention within each feature. Learned item-summary and feature-summary tokens connect these operations: item-summary tokens are concatenated along the feature axis, whereas feature-summary tokens are concatenated along the item axis. By repeating this exchange throughout the network, CAST progressively refines feature interactions using support-set context while retaining cell-level representations. This contrasts with staged architectures that aggregate feature embeddings into a single row representation before dataset-level in-context learning~\cite{qu2025tabicl,grinsztajn2026tabpfn3,qu2026tabiclv2}.

This report is the first public technical description of \EXAONETabular{}, and its main contributions are the following.

\begin{itemize}
    \item \textbf{State-of-the-art compact tabular foundation models.}
    \EXAONETabular{} provides separately trained classification and regression models with approximately 20.8M and 21.1M parameters, respectively, achieving state-of-the-art or near-state-of-the-art performance across four public benchmarks. On TabArena~\cite{erickson2025tabarena}, the classification model ranks first among all evaluated methods using a single fixed default configuration, without dataset-specific tuning or post-hoc ensembling, while the regression model reaches the performance regime of the 1.64B-parameter TabFM~\cite{google2026tabfm} at roughly 1/11 the inference cost. On BCCO~\cite{zhang2025limix} and TALENT~\cite{liu2024talent}, \EXAONETabular{} ranks second in classification and first in regression. On ScoringBench~\cite{landsgesell2026scoringbench}, it achieves the best mean rank for both point-estimation and predictive-distribution quality, leading the $R^2$, RMSE, and CRPS evaluations.

    \item \textbf{Cross-Axis Summary Transformer (CAST).}
    We introduce CAST, an interleaved feature--item architecture for tabular in-context learning. Each Transformer layer couples feature-axis processing within items with support-conditioned item-axis processing within features through distinct item-summary and feature-summary pathways. Unlike staged architectures that compress feature representations before the principal dataset-level ICL stage~\cite{qu2025tabicl,grinsztajn2026tabpfn3,qu2026tabiclv2}, CAST preserves cell-level representations throughout the hierarchy, allowing feature representations and support-set context to be progressively refined across layers.

    \item \textbf{Strong performance under realistic missing-data conditions.}
    \EXAONETabular{} consumes missing values natively, without requiring imputation, and incorporates missingness information when constructing feature representations. On BCCO~\cite{zhang2025limix}, which includes a substantial number of datasets with missing entries, the classification model ranks second overall while the regression model ranks first, demonstrating strong predictive performance on realistic incomplete tabular data.
\end{itemize}

The remainder of this report reviews related work, describes the architecture and synthetic prior in \cref{sec:modeling}, presents the training recipe in \cref{sec:training}, defines the evaluation protocol in \cref{sec:evaluation}, and discusses limitations and future work in \cref{sec:limitations}.

\subsection*{Related Work and Architectural Positioning}

\EXAONETabular{} follows the synthetic-prior, in-context-learning paradigm established by Prior-Fitted Networks~\cite{muller2022transformers} and instantiated for tabular prediction by the TabPFN line of work~\cite{hollmann2023tabpfn,hollmann2025accurate,grinsztajn2025tabpfn25}. TabPFN v1~\cite{hollmann2023tabpfn} represents each sample as a single token and performs in-context learning directly over sample embeddings. TabPFN v2~\cite{hollmann2025accurate} instead maintains cell-level representations and alternates attention along the feature and sample axes, while TabPFN-2.5~\cite{grinsztajn2025tabpfn25} extends this architectural line to substantially larger datasets and improved predictive performance.

A second family separates feature representation from dataset-level in-context learning by compressing each row into a fixed-dimensional embedding before a separate ICL stage. TabICL~\cite{qu2025tabicl} constructs distribution-aware feature representations, models within-row interactions, and aggregates them into one row embedding per sample prior to dataset-level ICL; TabICLv2~\cite{qu2026tabiclv2} extends this design, and TabPFN-3~\cite{grinsztajn2026tabpfn3} follows a similar multi-stage organization. TabFM~\cite{google2026tabfm}, a 1.64B-parameter model pretrained on hundreds of millions of SCM-generated synthetic tables, combines elements of both architectural families. According to its official technical release and model documentation---although no archival architecture paper has been published at the time of writing---TabFM performs repeated cell-level feature- and sample-axis attention, followed by row compression and a separate ICL Transformer. It therefore retains a terminal compression boundary between cell-level contextualization and row-level ICL.

Other recent tabular foundation models differ along the pretraining objective and context-construction axes. TabDPT~\cite{ma2024tabdpt} is pretrained on real tabular data with a masked-column objective and retrieves a bounded, query-relevant context rather than conditioning on the full support set. LimiX~\cite{zhang2025limix} retains SCM-based synthetic pretraining but reformulates prediction as masked joint-distribution modeling, expressing classification, regression, imputation, and generation as conditional queries to a single model.

\EXAONETabular{} occupies a different position in this design space. Unlike TabICL, TabICLv2, TabPFN-3, and TabFM, it does not impose a one-time compression boundary between feature processing and dataset-level contextualization. Instead, feature-axis processing and item-axis contextualization are interleaved throughout the Transformer hierarchy, allowing support-set context formed at one layer to influence feature representations at subsequent layers. Relative to the TabPFN v2 family, item-axis interaction augments each feature's support sequence with learned feature-summary tokens that persist across Transformer layers. This organization preserves recurrent feature--context interaction while maintaining cell-level states throughout both axis-wise operations.

\EXAONETabular{} further maintains item-summary and feature-summary tokens with explicit and complementary roles. Item-summary slots are instantiated for every item and joined to its feature-axis sequence, whereas feature-summary slots are instantiated for every feature and joined to its support item-axis sequence. These pathways preserve cell-level representations throughout the hierarchy rather than converging early into a single fixed \texttt{[CLS]}-style row representation processed by a separate ICL module. Together, these design choices constitute the Cross-Axis Summary Transformer (CAST), described in detail in Section~\ref{sec:architecture}. Table~\ref{tab:model-summary} summarizes the core components of \EXAONETabular{}.

\begin{table}[H]
\centering
\caption{Core components of \EXAONETabular{}.}
\label{tab:model-summary}
\small
\begin{tabularx}{\linewidth}{
@{}>{\raggedright\arraybackslash}p{1.82in}Y@{}
}
\toprule
\textbf{Component} & \textbf{Role} \\
\midrule

Cross-Axis Summary Transformer (CAST) &
Interleaves feature-axis processing within items with support-conditioned item-axis processing within features; item-summary tokens join feature-axis sequences, while feature-summary tokens join support item-axis sequences. \\

Synthetic SCM prior &
Generates causal, categorical, nonlinear, noisy, and missing-data prediction tasks without using real pretraining tables. \\

Task-specific prediction heads &
Uses separate classification and regression heads for the independently trained task-specific checkpoints. \\

Training and inference pipeline &
Applies task-specific optimization and inference procedures, including preprocessing, test-time ensembling, and chunked support-conditioned inference. \\

\bottomrule
\end{tabularx}
\end{table}

\section{Modeling}
\label{sec:modeling}

\EXAONETabular{} is LG AI Research's tabular foundation model family for classification and regression, comprising separately trained task-specific models that share the same core architecture and are pretrained entirely on synthetic data. This section specifies the model architecture (\S\ref{sec:architecture}), prediction heads (\S\ref{sec:heads}), SCM prior (\S\ref{sec:prior}), preprocessing and test-time ensembling (\S\ref{sec:preprocessing}), and inference procedure (\S\ref{sec:inference}).

\subsection{Model Architecture}
\label{sec:architecture}

\begin{figure}[t]
\centering
\includegraphics[width=0.98\linewidth]{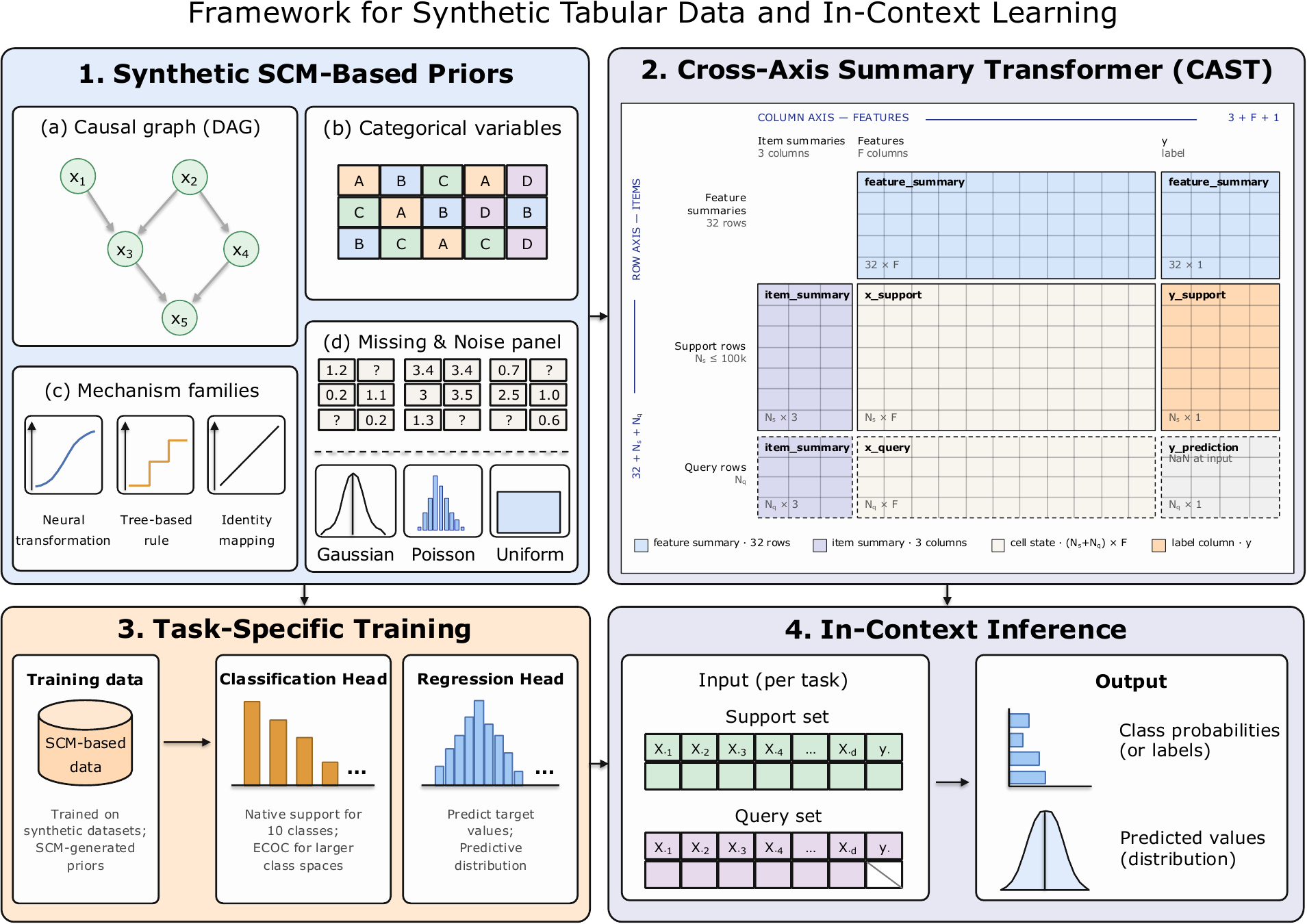}
\caption{Overall framework of \EXAONETabular{}. The model combines an SCM-based synthetic prior, a \code{Cross-Axis Summary Transformer} with interleaved feature-axis and item-axis attention, task-specific heads for classification and regression, and in-context inference on support and query sets. Item-summary tokens are concatenated with cell states along the feature axis, while feature-summary tokens are concatenated with support-item states along the item axis.}
\label{fig:exaone-architecture}
\end{figure}

An overview of \EXAONETabular{}'s architecture is shown in Figure~\ref{fig:exaone-architecture}. At the core of \EXAONETabular{} is the \emph{Cross-Axis Summary Transformer} (CAST), which jointly processes a table along its feature and item axes while retaining cell-level representations throughout the network. Each Transformer layer applies two feature-axis attention operations within every item, followed by support-conditioned item-axis interaction within every feature, and this pattern is repeated across all 12 layers.

\paragraph{Design Motivation.}
CAST is designed to keep feature processing and support-set contextualization coupled throughout the Transformer hierarchy. Cross-axis information is routed through two complementary summary states: item-summary tokens ($S_i$) and feature-summary tokens ($S_f$). Item-summary tokens are concatenated along the feature axis, whereas feature-summary tokens are concatenated along the item axis. This organization allows feature representations and support-set context to be progressively updated across layers without collapsing the feature axis into a fixed row representation.

\paragraph{Feature Encoding.}
Each input feature is encoded independently. Finite input values and indicators for missing or infinite values are jointly processed by a dedicated missing-value handling step. The model can therefore use both observed values and missingness patterns when constructing feature representations.

\paragraph{Feature-Axis Processing with Item-Summary Tokens.}
Before feature-axis self-attention, each cell reads the feature-summary states attached to its feature through cross-attention and incorporates the result through a residual connection. For every item, three item-summary tokens are then concatenated with its cell states along the feature axis. Self-attention over this combined sequence jointly updates the item-summary tokens and cell states. \EXAONETabular{} repeats this feature-axis operation twice per layer without collapsing the feature axis into a fixed item embedding, so cell-level representations remain available throughout the Transformer hierarchy.

\paragraph{Item-Axis Processing with Feature-Summary Tokens.}
Before item-axis attention, each cell reads the three item-summary states attached to its item through cross-attention and incorporates the result through a residual connection. For every feature, 32 feature-summary tokens are then concatenated with the support-item cell states along the item axis. The attention computation is divided into two paths:
\begin{itemize}
    \item \textbf{Support path:} the feature-summary tokens and support-item cell states are jointly updated through self-attention over their combined item-axis sequence.
    \item \textbf{Query path:} each query-item cell cross-attends to the same combined feature-summary and support-item input sequence.
\end{itemize}

Query items neither attend to one another nor update the feature-summary tokens or support-item states. The item-axis context available to every query is therefore determined exclusively by the labeled support set and the feature-summary states entering the layer. Consequently, the prediction for an individual query row can be written as
\[
p(y_j \mid x_j, D_{\mathrm{support}}),
\]
without dependence on other query examples in the same inference batch. This conditional independence enables query examples to be processed in separate chunks without changing the support-conditioned prediction, apart from minor numerical differences introduced by the underlying computation kernels.

In short, item-summary tokens are concatenated with each item's cells along the feature axis, whereas feature-summary tokens are concatenated with each feature's support-item cells along the item axis.

\paragraph{Scalable Attention Normalization.}
\EXAONETabular{} uses Scalable Softmax (SSMax), which adjusts attention normalization according to the number of context tokens. Standard softmax attention can become increasingly diffuse as context length grows; SSMax was introduced to counteract this effect and improve context-length generalization~\citep{nakanishi2025ssmax}.

\EXAONETabular{} uses a 192-dimensional embedding, 6 attention heads, and 12 Transformer layers. The classification checkpoint contains 20,807,866 parameters (approximately 20.81M), comprising approximately 20.65M backbone parameters and 0.156M task-head parameters. The regression checkpoint uses the same backbone together with a distributional regression head, resulting in approximately 21.11M parameters in total. Table~\ref{tab:model-config} summarizes the core architectural configuration.

\begin{table}[H]
\centering
\caption{Core model configuration.}
\label{tab:model-config}
\small
\begin{tabularx}{0.86\linewidth}{@{}Y>{\raggedleft\arraybackslash}p{2.2in}@{}}
\toprule
\textbf{Configuration} & \textbf{Value} \\
\midrule
Embedding dimension & 192 \\
Attention heads & 6 \\
Transformer layers & 12 \\
Feed-forward expansion & $4\times$ \\
Feature-attention operations per layer & 2 \\
Item-summary tokens per item & 3 \\
Feature-summary tokens per feature & 32 \\
Attention normalization & SSMax \\
Total parameters, classification & 20.81M \\
Total parameters, regression & 21.11M \\
\bottomrule
\end{tabularx}
\end{table}

\subsection{Prediction Heads}
\label{sec:heads}

\paragraph{Classification.}
The native classification head is a two-layer MLP decoder that maps the representation of each query row to logits over up to 10 classes, with class probabilities obtained through softmax. Datasets with more than 10 classes are handled at inference time through an Error-Correcting Output Code decomposition, described in Section~\ref{sec:inference}.

During test-time ensembling, class-label indices may be permuted to reduce sensitivity to the arbitrary numerical ordering of class labels.

\paragraph{Regression.}
The regression head directly predicts 999 conditional quantiles for each query item. Specifically, it outputs estimates
\[
\hat Q(\tau_k\mid x_j,D_{\mathrm{support}}),
\qquad
\tau_k=\frac{k}{1000},\quad k=1,\dots,999,
\]

For point prediction, the default inference procedure first sorts the predicted quantiles to guard against quantile crossing and then computes a trapezoidal average over the central 99.8\% of the quantile function. This provides an approximation to the conditional mean. The median, corresponding to the $\tau=0.5$ quantile, is also available as an alternative point-estimation readout.

\subsection{Synthetic SCM Prior}
\label{sec:prior}

\EXAONETabular{} is pretrained without real tabular datasets. Each training episode is generated from a synthetic structural causal model (SCM) prior designed to cover a broad range of tabular data-generating processes. The generator is organized into five stages as follows.


\paragraph{(i) Sample task hyperparameters.}
The generator first samples high-level task settings, including the number of rows, number of features, number of classes, graph size, component structure, and mechanism families. During pretraining, task size and structure are sampled over a broad range of support-set sizes, feature counts, class counts, and graph configurations, subject to a per-task compute budget.

\paragraph{(ii) Sample a DAG and node noise.}
Conditioned on the sampled task settings, the generator constructs a directed acyclic graph that defines dependencies among latent variables. The graph may contain one or two weakly connected components, and a randomized topology procedure diversifies the dependency structure across tasks. Each edge is assigned one of three mechanism families: a neural transformation, a tree-based rule, or an identity mapping. Independent noise is sampled per node and injected during SCM computation.

\paragraph{(iii) Compute the SCM in topological order.}
A topological ordering of the DAG determines the evaluation sequence. Root nodes are initialized from exogenous random draws, after which each remaining node is computed from its parents by a scaled sum of per-edge mechanism outputs. Neural mechanisms apply smooth nonlinear transformations, while tree-based mechanisms apply piecewise, threshold-like rules, with optional per-node Gaussian noise. This process produces fully evaluated SCMs containing diverse nonlinearities, interactions, and noise levels.

\paragraph{(iv) Choose observed features and target.}
After all SCM variables have been computed, a subset of nodes is selected as observed input features and another node is selected as the prediction target. The resulting rows are assembled into a tabular dataset and divided into labeled support examples and query examples. Classification and regression episodes use the same general generation pipeline while differing in target construction and task-specific sampling settings.

\paragraph{(v) Post-process and filter.}
The extracted table is transformed to reproduce common characteristics of real tabular data. Post-processing includes categorical discretization, nonlinear feature warping, feature-wise quantization, and task-specific missingness injection. An optional quality filter can screen out tasks that are too easy, too noisy, or insufficiently informative. The resulting dataset forms a synthetic training episode for in-context pretraining.

\subsection{Preprocessing and Test-Time Ensembling}
\label{sec:preprocessing}

The inference pipeline provides a scikit-learn-compatible interface through \code{fit} and \code{predict}. In this interface, \code{fit} registers and preprocesses the labeled support set; it does not update the pretrained model parameters.

When the number of support rows exceeds the configured inference limit, the support set is subsampled before preprocessing. At inference time, predictions can be ensembled across estimators constructed with different preprocessing and permutation configurations. Candidate feature transformations include raw features, robust scaling, power transformations, and quantile transformations. Categorical encodings, feature orderings, and class-label indices may also be permuted.

Both classification and regression support test-time ensembling. For regression, predictions from individual estimators are aggregated across preprocessing and permutation configurations. The default number of test-time estimators is eight for classification, and the ensemble size can be adjusted according to latency and accuracy requirements.

\subsection{Inference}
\label{sec:inference}

The complete inference pipeline may combine multiple preprocessing and permutation estimators, query chunks, and, for classification tasks beyond the native class limit, multiple ECOC subproblems. A complete prediction request may therefore involve multiple model forward operations, although no dataset-specific parameter updates are performed.

\paragraph{Error-Correcting Output Code Decomposition.}
For classification datasets with more than 10 classes, the inference pipeline uses an Error-Correcting Output Code (ECOC) decomposition~\citep{dietterich1995ecoc} to extend the fixed-size classification head to larger label spaces. Each row of the code matrix relabels the original classes into an alphabet of up to 10 symbols, defining multiclass classification subproblems. All subproblems are evaluated using the same pretrained model, and their output probabilities are decoded into probabilities over the original classes using the code matrix.

\paragraph{Inference Efficiency and Chunking.}
Inference can be organized along the estimator, ECOC-subproblem, and query dimensions. Estimators are independent because they may apply different transformations or permutations to the support set, while ECOC subproblems use different support-label assignments. Consequently, support-conditioned states generally cannot be shared across estimators or ECOC subproblems.

For a fixed estimator and ECOC subproblem, the support set remains unchanged across query chunks. The model can therefore reuse support-side keys and values when such reuse is compatible with the feature encoding applied to the chunks. The inference executor automatically selects between cached and uncached execution according to the input configuration, device, and available memory. With cached execution, the support-side keys and values are constructed once and reused across query chunks. With uncached execution, each query chunk is evaluated together with the support set and the corresponding support-side values are recomputed.

The attention implementation dispatches between FlashAttention~\citep{dao2022flashattention} and Memory-Efficient Attention~\citep{rabe2021selfattention} according to the sequence length and tensor configuration. The default compute dtype is fp16, while query-axis and estimator-axis chunking are used to control peak memory consumption for larger support and query sets.

\paragraph{Cost Structure and Support Recomputation.}
For a single estimator and ECOC subproblem, the computation is dominated by three operations. Feature-axis attention is linear in the number of items and quadratic in the number of feature groups. Item-axis support self-attention scales approximately quadratically with the support size $N_s$, while query-to-support cross-attention scales with the product of the query size $N_q$ and the support size. The leading item-axis cost is therefore approximately
\[
\mathcal{O}\!\left(N_s^2 + N_sN_q\right)
\]
when the support context is constructed once.

The number of distinct support-conditioned states required for one prediction request is
\[
E \cdot R,
\]
where $E$ is the number of ensemble estimators and $R$ is the number of ECOC subproblems, with $R=1$ when ECOC decomposition is not required. These factors are inherent because each estimator may use a differently transformed support set and each ECOC subproblem changes the support-label assignment.

If the query set is divided into $n_{qc}$ chunks, the number of estimator--subproblem--query-chunk evaluations is
\[
E \cdot R \cdot n_{qc}.
\]
When cached execution is selected, the support-side keys and values are constructed
\[
E \cdot R
\]
times and reused across query chunks. Under uncached execution, they are instead constructed
\[
E \cdot R \cdot n_{qc}
\]
times. Thus, query chunking preserves the total query-to-support attention cost up to chunking overhead, but uncached execution additionally repeats the support-side construction for each chunk. The value of $n_{qc}$ and the use of caching are determined at inference time rather than fixed by the default configuration.

\section{Training}
\label{sec:training}

Classification and regression share the main architecture but use separately configured training recipes. Training is performed entirely on synthetically generated tables using bf16 precision. Across the training lineage used for the released models, classification processes approximately 30 million synthetic table instances, while regression processes approximately 10 million. Here, the number of tables refers to the total number of synthetic table instances processed during optimization.

\subsection{Optimization}

Selected matrix-valued parameters, including most attention and feed-forward weight matrices, are optimized with Muon~\citep{liu2025muon}, which applies a momentum-based update followed by matrix orthogonalization. The remaining parameters are optimized with AdamW~\citep{loshchilov2019adamw}. The optimizer assignments are therefore
\begin{equation}
\theta_{\mathrm{matrix}}\leftarrow\mathrm{Muon},
\qquad
\theta_{\mathrm{remaining}}\leftarrow\mathrm{AdamW}.
\end{equation}

Classification learning rates are $1\times10^{-4}$ for AdamW parameters and $2\times10^{-4}$ for Muon parameters. For regression, the corresponding learning rates are $3.8\times10^{-5}$ and $2\times10^{-4}$, respectively. Learning rates and weight decay are adjusted in subsequent training stages.

\subsection{Learning-Rate Schedule}

Training primarily uses Warmup-Stable-Decay (WSD) learning-rate schedules~\citep{wen2024wsd}. WSD consists of an initial warmup phase, an extended stable phase, and a terminal decay phase. Additional continuation and adaptation training uses closely related cosine or constant-learning-rate schedules depending on the training objective.

\paragraph{Exponential Moving Average}

Both classification and regression maintain EMA model weights throughout training. For parameters $\theta_t$ at optimization step $t$, the EMA parameters are updated as
\begin{equation}
\theta_t^{\mathrm{EMA}}
=
\gamma\theta_{t-1}^{\mathrm{EMA}}
+
(1-\gamma)\theta_t.
\end{equation}
A decay of $\gamma=0.999$ is used for most training, with stage-specific adjustments in later classification training. Evaluation and inference use the EMA weights.

\section{Evaluation}
\label{sec:evaluation}

\subsection{Benchmarks}

We evaluate EXAONE Tabular on four public benchmarks: TabArena~\citep{erickson2025tabarena}, BCCO (Balanced Comprehensive Challenging Omni-domain)~\citep{zhang2025limix}, TALENT~\citep{liu2024talent}, and ScoringBench~\citep{landsgesell2026scoringbench}. The four benchmarks cover complementary evaluation regimes and are reported separately rather than pooled into a single heterogeneous collection. TabArena enables a broad comparison with recent tabular foundation models and established task-specific methods. BCCO complements this evaluation by including datasets with missing values, thereby reflecting more realistic tabular data conditions. TALENT further broadens the evaluation with a substantially larger collection of classification and regression datasets spanning diverse tabular tasks. ScoringBench adds a regression-only evaluation that assesses both conventional point estimates and full predictive distributions through proper scoring rules.

For classification, TabArena contains 30 binary and 8 multiclass datasets, while BCCO contains 71 binary and 35 multiclass datasets. TALENT contains 200 classification datasets, including 120 binary and 80 multiclass datasets. To ensure a common comparison set across models with different class-count limits, we exclude 12 TALENT datasets with more than 10 target classes. BCCO additionally contains 50 regression datasets, while TALENT contains 100 regression datasets. ScoringBench is continuously maintained: its original release reports 97 regression datasets, while the live configuration accessed in August 2026 enumerates 104 candidate datasets. Because dataset validation, model-result coverage filtering, and complete-case selection are applied separately for each metric, the effective number of datasets used in a leaderboard comparison may be smaller and can differ across metrics.

On TabArena, EXAONE Tabular is compared with 13 baselines spanning recent tabular foundation models, gradient-boosted decision trees, neural tabular architectures, and AutoML systems. For BCCO and TALENT, we conduct a controlled comparison against six tabular foundation models: TabSwift~\citep{liu2026tabswift}, TabDPT~\citep{ma2024tabdpt}, TabPFN-3~\citep{grinsztajn2026tabpfn3}, TabICLv2~\citep{qu2026tabiclv2}, LimiX~\citep{zhang2025limix}, and TabFM~\citep{google2026tabfm}. For ScoringBench, we compare against all methods eligible for each metric on the official live leaderboard at the time of access. Its comparison set is therefore metric-dependent, particularly for distributional metrics that require a full predictive distribution.

\subsection{Evaluation Protocol}
\paragraph{TabArena Protocol} For TabArena, we use the results reported on the official leaderboard rather than re-running each comparison model. \EXAONETabular{} is evaluated using the same datasets, splits, metrics, and aggregation protocol to ensure direct comparability with the leaderboard results. The comparison covers the full set of leaderboard methods, spanning pretrained tabular foundation models, tree-based methods, neural networks, and AutoML systems; baseline-specific preprocessing, hyperparameter optimization, and ensembling therefore follow the TabArena evaluation protocol. Classification and regression performance are measured with the per-task TabArena metrics and aggregated into Elo ratings following the same protocol. All TabArena Elo values, confidence intervals, and timing measurements reported here (Figures~\ref{fig:size-performance-pareto} and \ref{fig:pareto_infer_sidebyside}--\ref{fig:pairwise-winrate}) are taken from the official leaderboard (accessed August 2026).


\paragraph{BCCO and TALENT Protocol}
For BCCO and TALENT, all comparison models were evaluated using their official GitHub repositories and publicly released checkpoints. We preserved each model’s official preprocessing and inference procedures whenever possible. To control computational cost across datasets of different scales, the training or in-context support set was capped at 50,000 samples per fold for both classification and regression. When the training split exceeded this limit, 50,000 samples were drawn without replacement, while the validation and test splits remained unchanged.

All data-dependent preprocessing was fitted exclusively on the training split and then applied unchanged to the validation and test splits. For models with native categorical-feature support, we retained the default preprocessing pipeline. Models requiring numerical inputs used ordinal mappings fitted on the training data, with categories unseen at test time treated as missing values. When the number of features exceeded a model’s input limit, the input dimensionality was reduced according to its official implementation.

For classification, the same set of datasets was used across all comparison models. In TALENT, 12 datasets with more than 10 target classes were excluded because some of the evaluated tabular foundation models do not support larger numbers of classes.

\paragraph{ScoringBench Protocol}
For ScoringBench, we use the results reported on the official live leaderboard. The benchmark evaluates probabilistic regression using five-fold cross-validation with a fixed random seed and at most 3,000 observations per dataset. Scores are first averaged across folds for each model and dataset, and models are then compared by their mean dataset-level ranks. Point-estimation quality is measured using $R^2$ and RMSE, while CRPS evaluates the complete predictive distribution rather than only its mean. The leaderboard applies metric-specific coverage filtering, so the eligible models and effective dataset set may differ across metrics. All ScoringBench rankings reported in this report are taken from the official leaderboard (accessed August 2026).

\subsection{TabArena Results}

\begin{figure}[t]
\centering
\begin{subfigure}[b]{0.48\linewidth}
    \centering
    \includegraphics[width=\linewidth]{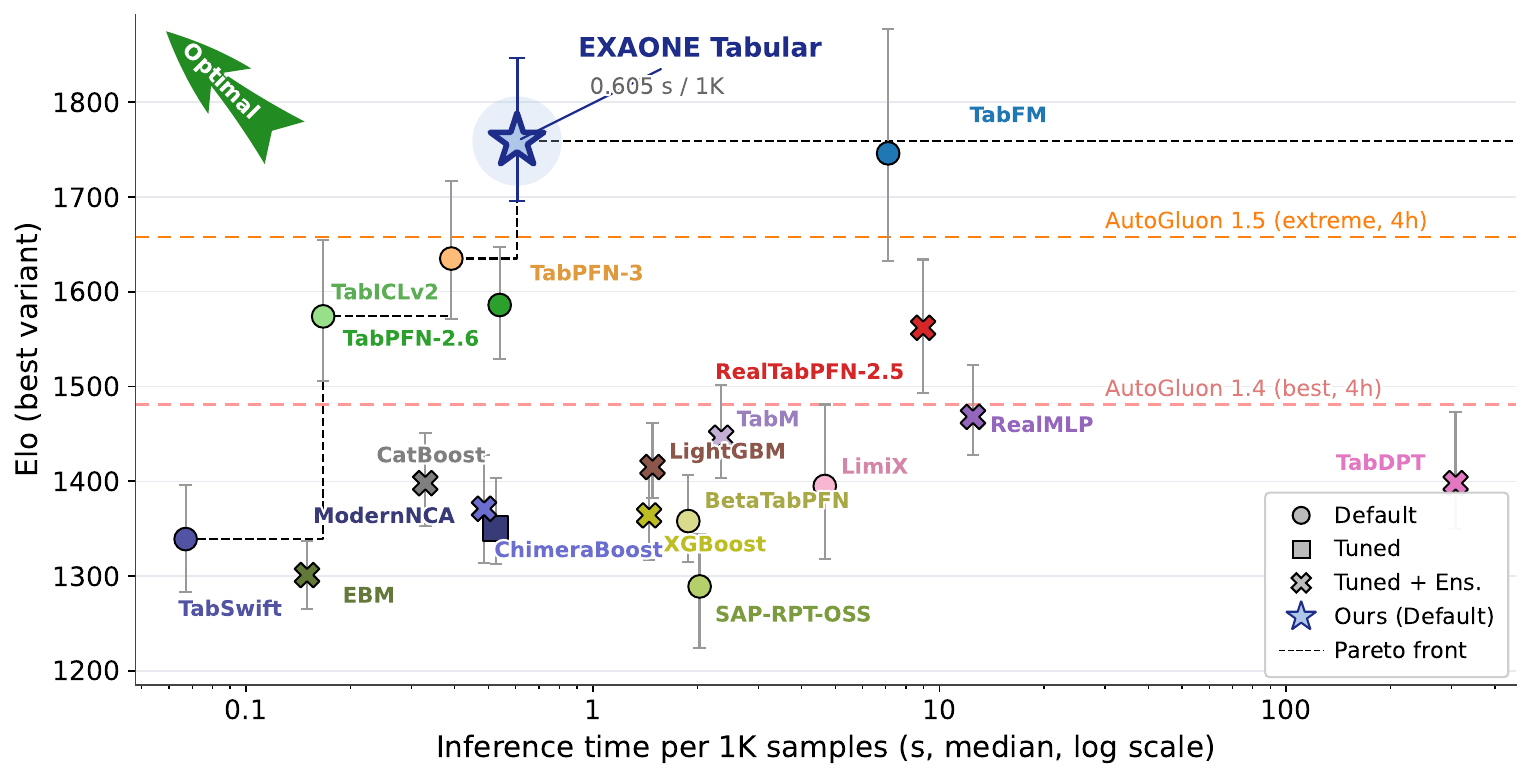}
    \captionsetup{justification=centering, singlelinecheck=off}
    \caption{Classification task}
    \label{fig:pareto_infer_cls}
\end{subfigure}
\hfill
\begin{subfigure}[b]{0.48\linewidth}
    \centering
    \includegraphics[width=\linewidth]{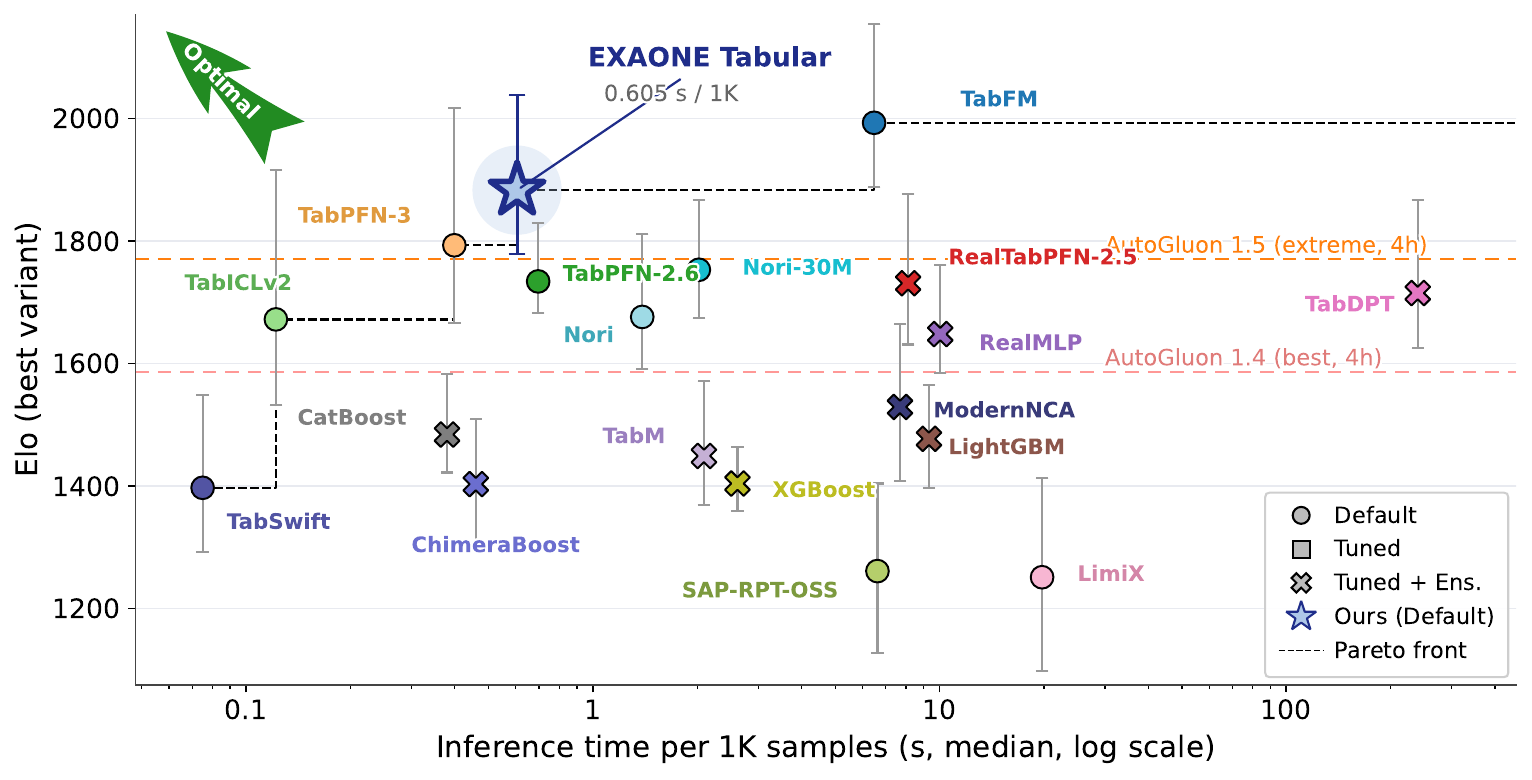}
    \captionsetup{justification=centering, singlelinecheck=off}
    \caption{Regression task}
    \label{fig:pareto_infer_reg}
\end{subfigure}

\caption{\textbf{Accuracy--latency Pareto fronts on TabArena for (a)~classification and (b)~regression tasks.} 
 Elo is plotted against median inference time per 1,000 samples (log
    scale); dashed staircases trace the Pareto fronts, and horizontal
    lines mark the 4-hour AutoGluon reference pipelines~\citep{erickson2020autogluon}. Using its default
    single-forward-pass inference at 0.605~s per 1,000 samples,
    \EXAONETabular{} lies on the Pareto front of both task types. On
    classification it attains the highest Elo on the entire leaderboard,
    exceeding TabPFN-3 by $\sim$125 Elo at comparable latency. On
    regression it reaches the accuracy regime of TabFM --- within
    overlapping 95\% confidence intervals --- at roughly $1/11$ the
    inference cost, and exceeds the regression-specialized Nori models~\citep{synthefy2026nori} by
    $\sim$140 Elo.}
\label{fig:pareto_infer_sidebyside}
\end{figure}
\begin{figure}[!t]
\centering
\begin{subfigure}[b]{0.48\linewidth}
    \centering
    \includegraphics[width=\linewidth]{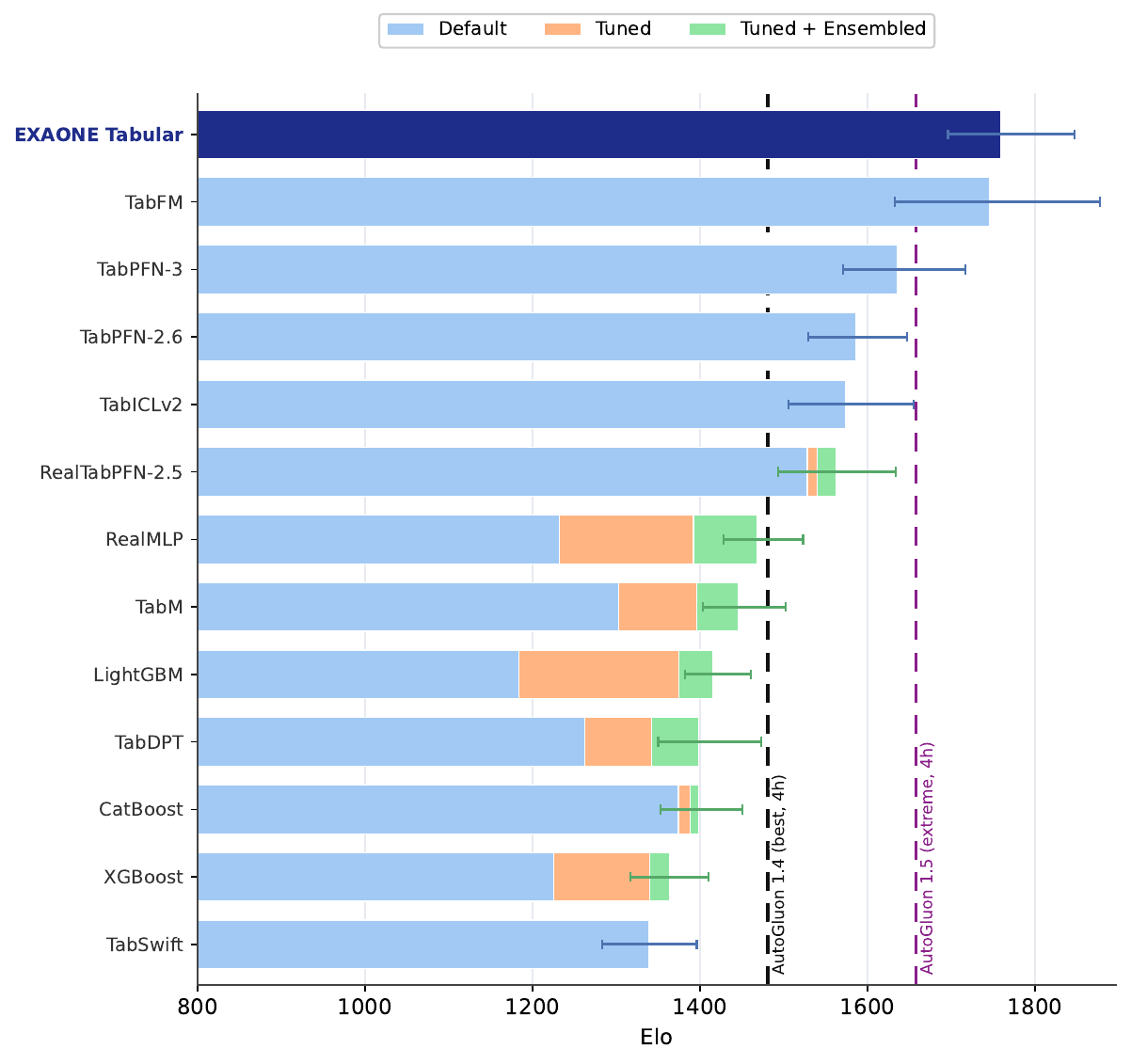}
    \captionsetup{justification=centering, singlelinecheck=off}
    \caption{Classification task}
    \label{fig:tuning_cls}
\end{subfigure}
\hfill
\begin{subfigure}[b]{0.48\linewidth}
    \centering
    \includegraphics[width=\linewidth]{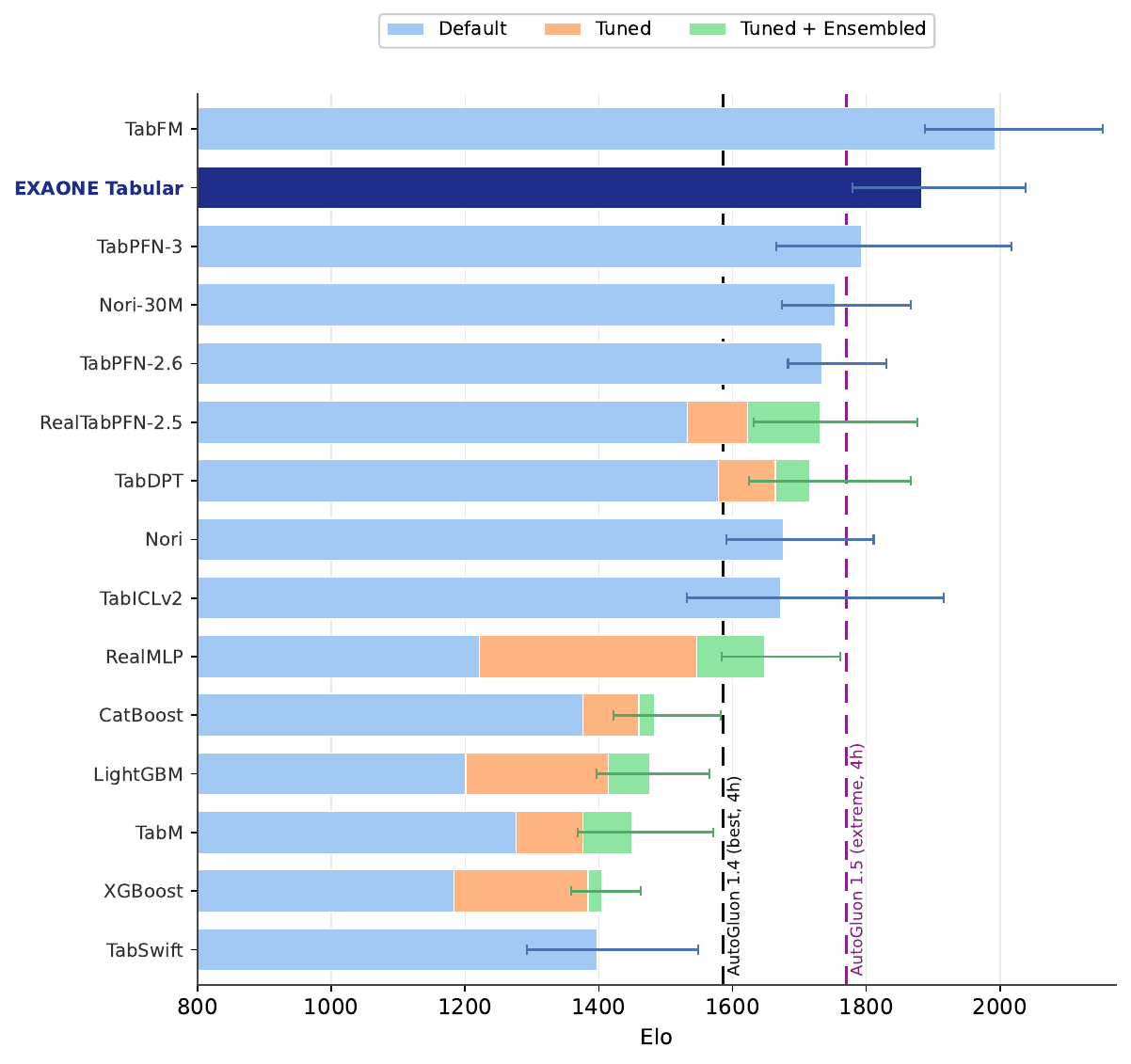}
    \captionsetup{justification=centering, singlelinecheck=off}
    \caption{Regression task}
    \label{fig:tuning_reg}
\end{subfigure}
\caption{
\textbf{Default vs.\ tuned and post-hoc ensembled TabArena Elo for
(a)~classification and (b)~regression.}
    Bars show each method's default configuration, with additional
    segments marking the gains from per-task tuning and post-hoc
    ensembling where those variants exist. Conventional GBDT and neural
    baselines depend heavily on both, worth $\sim$230 Elo for
    LightGBM~\citep{ke2017lightgbm} and RealMLP~\citep{holzmuller2024realmlp}
    on classification and over 420 Elo for RealMLP on regression
    alone. \EXAONETabular{} (navy), like most tabular foundation models,
    is evaluated in a single untuned default configuration, and ranks
    first on classification and, on regression, second only to TabFM ---
    a model $\sim$78$\times$ its size. Dashed lines mark the 4-hour
    AutoGluon~1.4/1.5 pipelines, both of which it clears on both task
    types.}
\label{fig:tuning_impact_sidebyside}
\end{figure}

TabArena is a continuously maintained benchmark for tabular machine learning that provides curated datasets, standardized evaluation pipelines, and strong implementations of classical, neural, and foundation-model baselines \citep{erickson2025tabarena}. TabArena comprises 51 real-world predictive tasks---38 classification and 13 regression tasks---manually selected from 1,053 candidate datasets. The benchmark covers IID datasets with 500 to 250,000 training samples and substantial variation in feature dimensionality and categorical-feature prevalence, providing a diverse test bed for tabular learning methods.

\begin{figure}[!t]
    \centering
    \includegraphics[width=0.6\columnwidth]
    {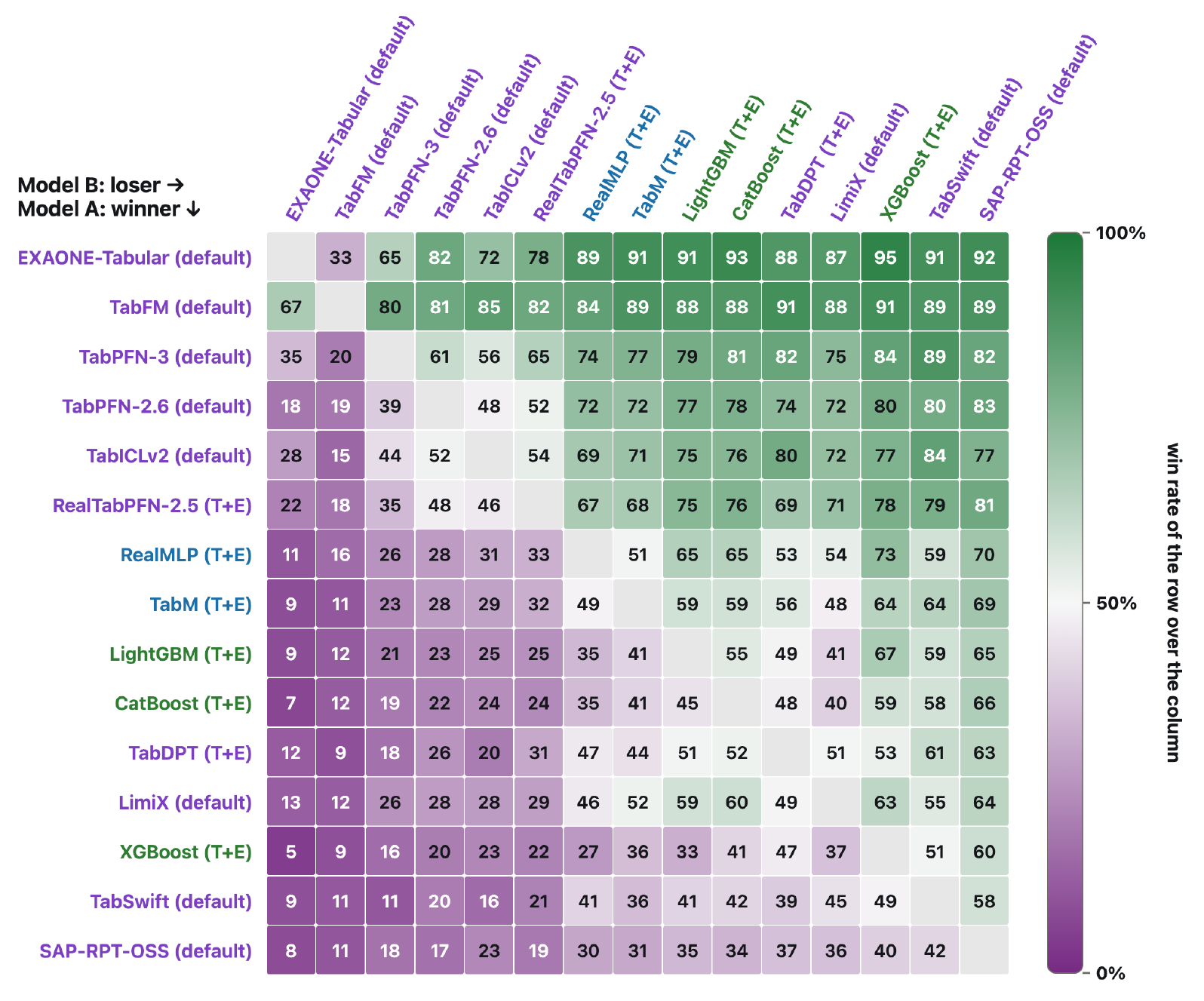}
\caption{\textbf{Pairwise win rates on the full TabArena benchmark (classification and regression).} Each entry $(A, B)$ reports the percentage of evaluation splits on which method $A$ attains a lower test error than method $B$; cells are shaded from purple (row loses) through white (parity) to green (row wins). Method labels are colored by family --- purple: tabular foundation models; blue: neural networks; green: GBDTs. With a single fixed default configuration, \EXAONETabular{} wins the majority of splits against the strongest variant of every competing method except TabFM: 65--92\% against the other tabular foundation models, and at least 89\% against every tuned-and-ensembled GBDT and neural-network baseline. The advantage is therefore consistent at the split level rather than being driven by a small subset of datasets.}
    \label{fig:pairwise-winrate}
\end{figure}

For the accuracy--latency comparison, each method is represented by its best-performing non-imputed variant on the leaderboard, with marker styles distinguishing default, tuned, and tuned-and-ensembled configurations; inference times correspond to the plotted variant. \EXAONETabular{} is evaluated with its single default configuration, refit on the full outer-training set, while baseline inference times are taken from the measurements released by TabArena, so its untuned default is compared against the strongest available configuration of every baseline. Figure~\ref{fig:pareto_infer_sidebyside} places these configurations on the accuracy--latency plane, separately for classification and regression. Using its default single-forward-pass inference at 0.605~s per 1,000 samples, \EXAONETabular{} lies on the Pareto front of both task types: on classification it attains the highest Elo on the entire leaderboard, outperforming TabPFN-3 by $\sim$125 Elo at comparable latency, and on regression it reaches the accuracy regime of TabFM --- the strongest evaluated model --- at roughly $1/11$ the inference cost. TabFM requires approximately 7.0~s per 1,000 samples, an $\sim$$11.5\times$ increase over \EXAONETabular{}.

Figure~\ref{fig:tuning_impact_sidebyside} contrasts the value of inference-time tuning and post-hoc ensembling across methods. Conventional GBDT and neural-network baselines climb the leaderboard only through hours of per-task tuning and post-hoc ensembling, whereas most tabular foundation models, including \EXAONETabular{}, are evaluated in a single untuned default configuration. Even within this default-only setting, \EXAONETabular{} reaches the top tier of both task types --- and the converse also holds: however large the tuning and ensembling gains, they never close the gap, as every tuned-and-ensembled variant on the leaderboard still falls short of its single untuned configuration on both tasks.

The pairwise analysis in Figure~\ref{fig:pairwise-winrate} complements the aggregate Elo ranking by measuring head-to-head consistency across evaluation splits. Elo limits the influence of unusually large performance margins by reducing each comparison to a win, tie, or loss, but a single aggregate rating can still mask intransitive reversals in which a lower-ranked method prevails in direct comparison. The matrix provides no evidence of such reversals for \EXAONETabular{}: no method ranked below it wins the majority of splits against it, and its win rates rise broadly with the Elo gap to each opponent. The consistently high win rates also indicate that its advantage is observed across the benchmark rather than being attributable to a small subset of evaluation splits.

\subsection{BCCO Results}
\label{sec:BCCO-benchmarks}
\begin{figure}[t]
\centering
\begin{subfigure}[b]{0.48\linewidth}
    \centering
    \includegraphics[width=\linewidth]{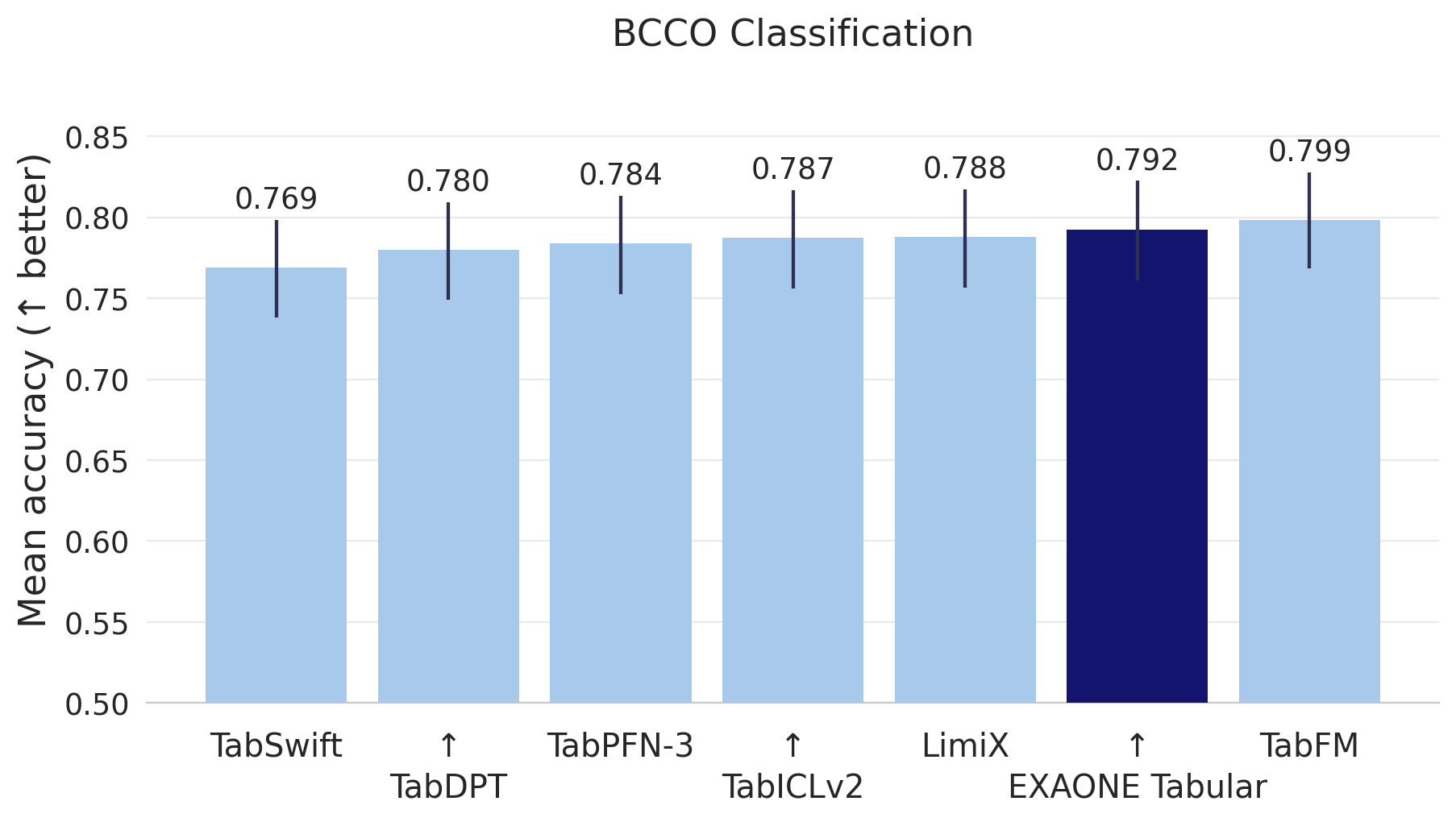}
    \caption{Classification performance on the BCCO benchmark, measured by mean test accuracy across 106 datasets.}
    \label{fig:bcco-cls-evaluation}
\end{subfigure}
\hfill
\begin{subfigure}[b]{0.48\linewidth}
    \centering
    \includegraphics[width=\linewidth]{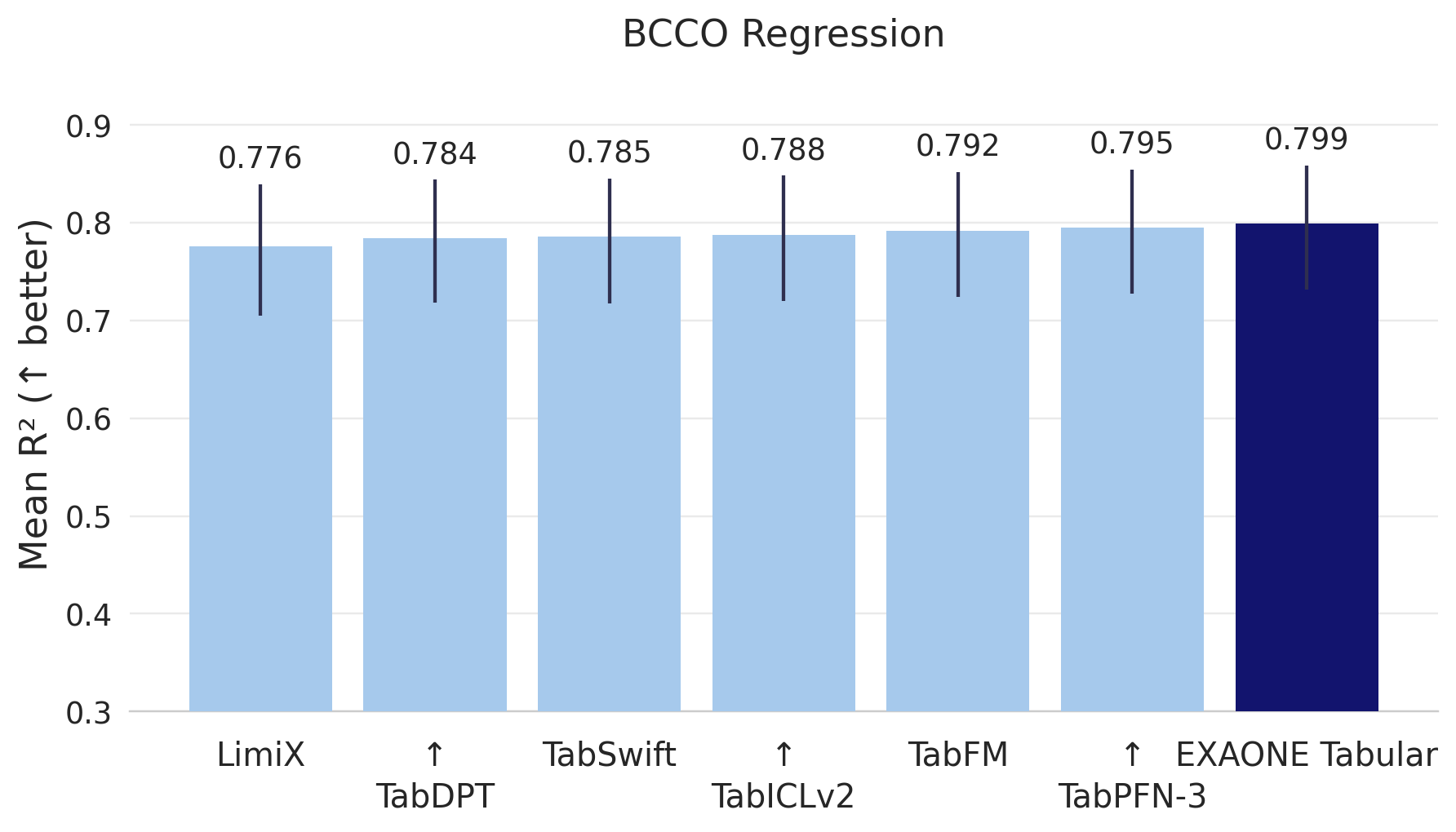}
    \caption{Regression performance on the BCCO benchmark, measured by mean $R^2$ across 50 datasets.}
    \label{fig:bcco-reg-evaluation}
\end{subfigure}
\caption{Performance comparison on the BCCO benchmark. \EXAONETabular{} achieves the second-highest mean accuracy on classification and the highest mean $R^2$ on regression, demonstrating competitive performance across both tasks. Error bars represent 95\% bootstrap confidence intervals.}
\label{fig:bcco-results}
\end{figure}

\paragraph{Classification}
Figure~\ref{fig:bcco-cls-evaluation} reports the mean test accuracy of the seven evaluated models across the 106 BCCO classification datasets. TabFM achieves the highest mean accuracy of 0.799, followed closely by \EXAONETabular{} with 0.792. \EXAONETabular{} outperforms LimiX, TabICLv2, TabPFN-3, TabDPT, and TabSwift, which achieve mean accuracies of 0.788, 0.787, 0.784, 0.780, and 0.769, respectively, while the absolute accuracy gap to TabFM is only 0.007.

This comparison highlights the favorable performance-to-parameter trade-off of \EXAONETabular{}. With 20.81M parameters, it is smaller than TabPFN-3 and TabDPT, comparable in scale to TabICLv2, and substantially smaller than the 1.64B-parameter TabFM. Notably, TabFM is evaluated with 32 ensemble members, whereas \EXAONETabular{} and the other comparison models use 8. Despite the larger model size and ensemble budget of TabFM, \EXAONETabular{} remains within 0.007 in mean accuracy. Although LimiX and TabSwift use fewer parameters, they achieve lower mean accuracy. These results indicate that \EXAONETabular{} provides a favorable balance between predictive performance, model size, and inference complexity.

The results also show that \EXAONETabular{} performs consistently across the diverse BCCO datasets. Since BCCO includes a substantial number of datasets containing NaN values, its second-best mean accuracy suggests that the model remains competitive under realistic incomplete-data conditions. 

Error bars represent 95\% bootstrap confidence intervals obtained by resampling datasets 10,000 times. As the confidence intervals of TabFM and \EXAONETabular{} overlap, the observed difference should not be interpreted as statistically significant without an additional hypothesis test. Nevertheless, \EXAONETabular{} achieves performance close to TabFM while using only approximately 1.3\% of its parameters and one-quarter of its ensemble members, demonstrating a strong balance between predictive performance and inference efficiency.

\paragraph{Regression}

Figure~\ref{fig:bcco-reg-evaluation} reports the mean $R^2$ of the seven evaluated models across the 50 BCCO regression datasets. \EXAONETabular{} achieves the highest mean $R^2$ of 0.799, followed by TabPFN-3 with 0.795. TabFM and TabICLv2 achieve 0.792 and 0.788, respectively, while TabSwift, TabDPT, and LimiX obtain 0.785, 0.784, and 0.776.

These results show that \EXAONETabular{} remains competitive across both classification and regression tasks on BCCO. In particular, \EXAONETabular{} ranks first in mean $R^2$ on regression, complementing its strong classification performance and demonstrating consistent predictive capability across diverse real-world tabular tasks.

\subsection{TALENT Results}
\label{sec:talent-benchmarks}

\begin{figure}[t]
\centering
\begin{subfigure}[b]{0.48\linewidth}
    \centering
    \includegraphics[width=\linewidth]{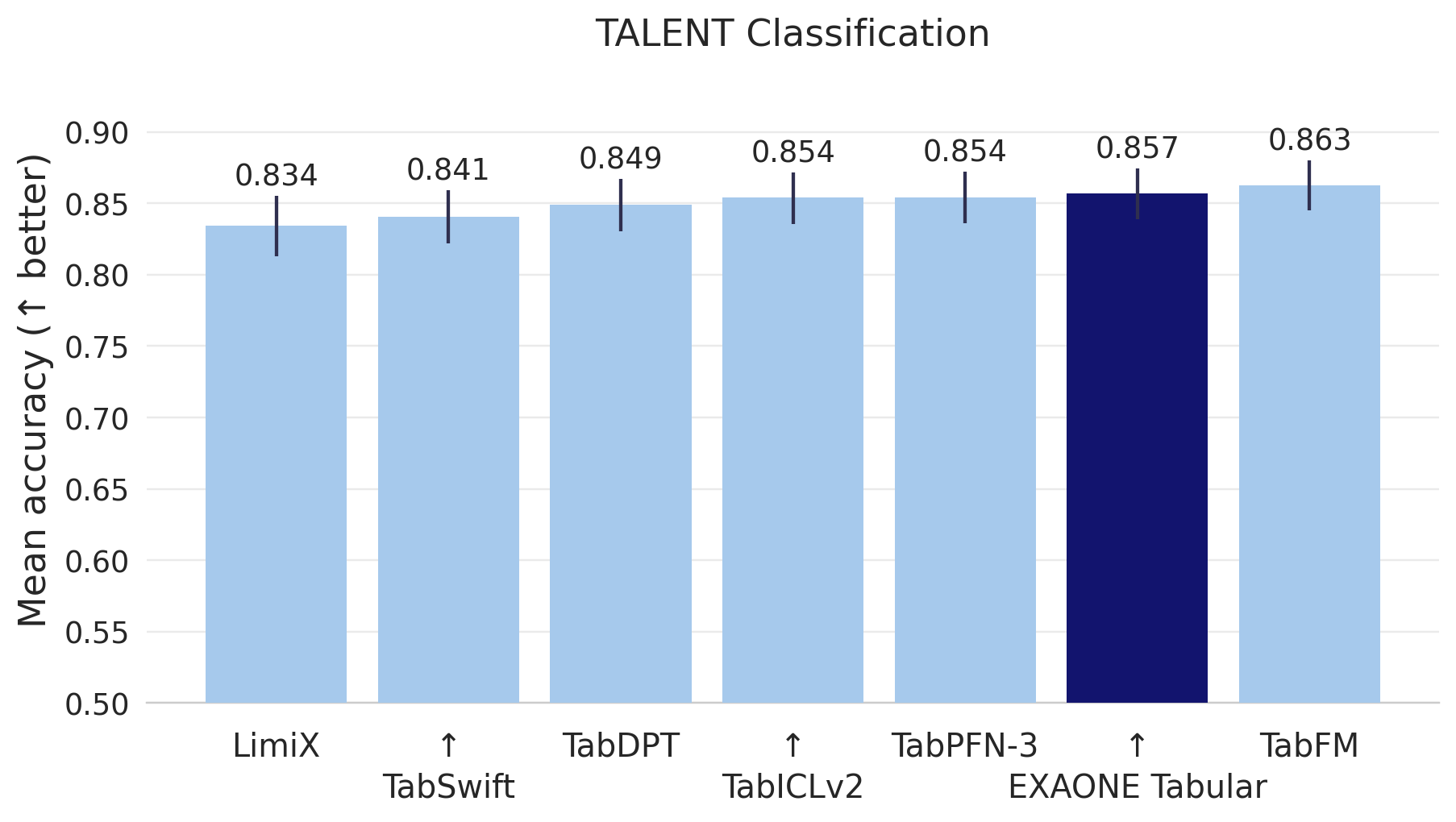}
    \caption{Classification performance on the TALENT benchmark, measured by mean test accuracy across 188 datasets.}
    \label{fig:talent-cls-evaluation}
\end{subfigure}
\hfill
\begin{subfigure}[b]{0.48\linewidth}
    \centering
    \includegraphics[width=\linewidth]{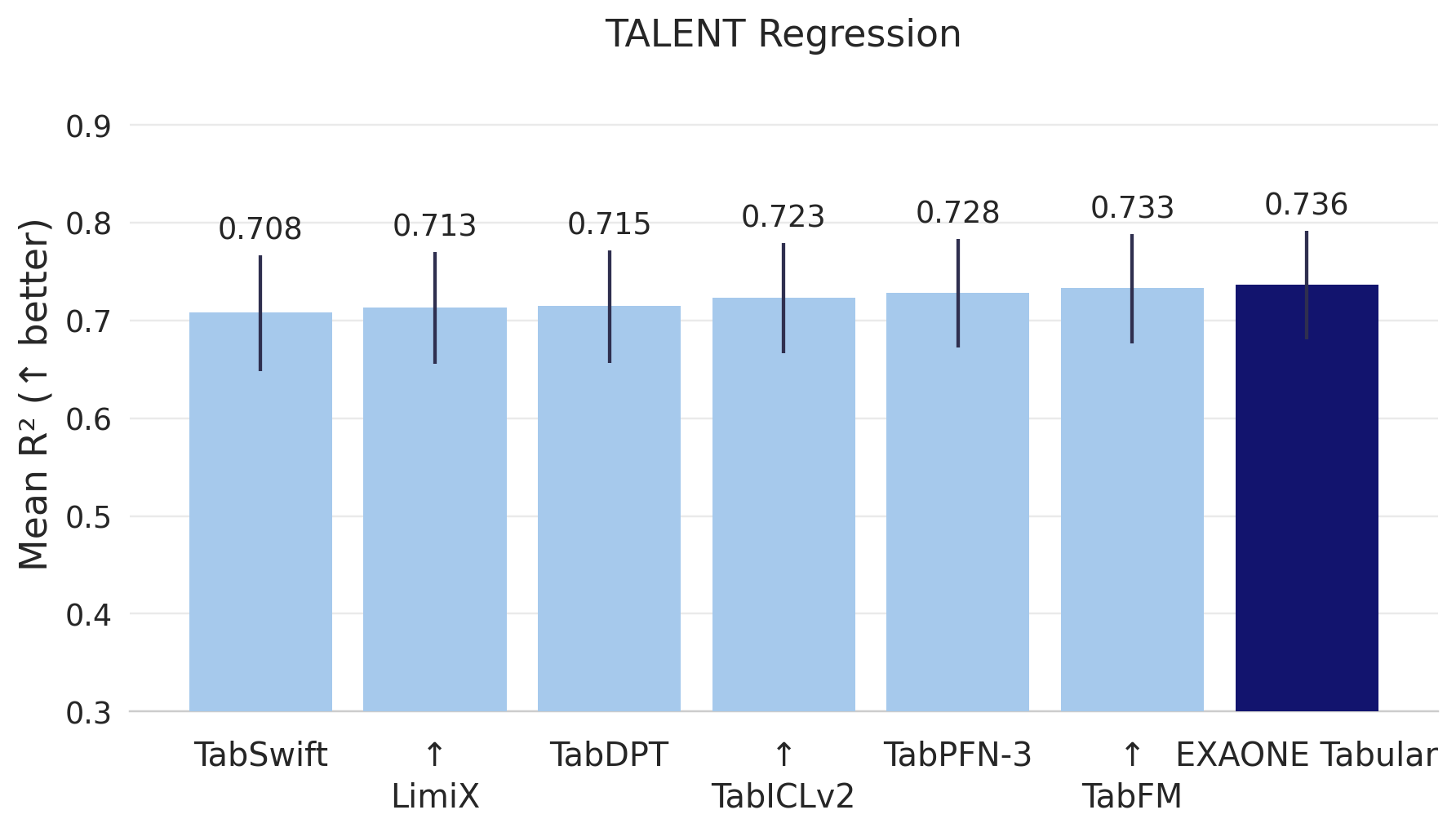}
    \caption{Regression performance on the TALENT benchmark, measured by mean $R^2$ across 100 datasets.}
    \label{fig:talent-reg-evaluation}
\end{subfigure}
\caption{Performance comparison on the TALENT benchmark. \EXAONETabular{} achieves the second-highest mean accuracy on classification and the highest mean $R^2$ on regression, demonstrating consistently competitive performance across both tasks. Error bars represent 95\% bootstrap confidence intervals.}
\label{fig:talent-results}
\end{figure}





\paragraph{Classification}
Figure~\ref{fig:talent-cls-evaluation} reports the mean test accuracy of the seven evaluated models across 188 TALENT classification datasets. We exclude 12 of the 200 classification datasets with more than 10 classes to ensure a common comparison set across models with different class-count limits. TabFM achieves the highest mean accuracy of 0.863, followed by \EXAONETabular{} with 0.857. TabPFN-3 and TabICLv2 both achieve 0.854, while TabDPT, TabSwift, and LimiX obtain 0.849, 0.841, and 0.834, respectively.

Although TabFM achieves the best overall result, the absolute accuracy gap to \EXAONETabular{} is only 0.006. \EXAONETabular{} outperforms all remaining tabular foundation models, demonstrating strong and consistent classification performance across the large and diverse TALENT benchmark.

Together with the results on TabArena and BCCO, the TALENT evaluation further shows that \EXAONETabular{} remains competitive across independently constructed public benchmarks with different dataset compositions and evaluation regimes. Its strong performance on TALENT provides additional evidence of robust cross-benchmark generalization.

\paragraph{Regression}
Figure~\ref{fig:talent-reg-evaluation} reports the mean $R^2$ of the seven evaluated models across the 100 TALENT regression datasets. \EXAONETabular{} achieves the highest mean $R^2$ of 0.736, followed by TabFM and TabPFN-3 with 0.733 and 0.728, respectively. TabICLv2, TabDPT, LimiX, and TabSwift obtain 0.723, 0.715, 0.713, and 0.708, respectively.

These results show that \EXAONETabular{} maintains strong regression performance across the large and diverse TALENT benchmark. Together with its second-best classification performance, the leading mean $R^2$ demonstrates that \EXAONETabular{} remains consistently competitive across both classification and regression tasks. Combined with the results on TabArena and BCCO, the TALENT evaluation further supports the model’s ability to generalize across independently constructed public benchmark suites.

\subsection{ScoringBench Results}
\label{sec:scoringbench-benchmarks}

\begin{figure}[t]
    \centering
    \adjustbox{
        max width=.94\linewidth,
        max totalheight=.43\textheight,
        keepaspectratio
    }{%
        \begingroup\endlinechar=-1%
\fontencoding{T1}\fontfamily{DejaVuSans-TLF}\selectfont
\contourlength{.95bp}
\def\PageW{720}
\def\PageH{430}
\def\PlotL{76}
\def\PlotR{678}
\def\PlotT{50}
\def\PlotB{365}
\def\MaxRank{50}
\definecolor{gridc}{RGB}{231,235,241}
\definecolor{spine}{RGB}{55,63,75}
\definecolor{secondary}{RGB}{101,111,126}
\definecolor{pointfill}{RGB}{93,153,207}
\definecolor{pointedge}{RGB}{46,104,157}
\definecolor{corefill}{RGB}{40,111,174}
\definecolor{coreedge}{RGB}{21,73,123}
\definecolor{oursfill}{RGB}{155,194,229}
\definecolor{oursedge}{RGB}{16,44,107}
\definecolor{frontier}{RGB}{94,200,224}
\definecolor{diagc}{RGB}{137,145,158}
\definecolor{tagborder}{RGB}{178,188,201}
\definecolor{panelbg}{RGB}{249,251,253}
\newcommand{\AxisText}[1]{\fontsize{12.8bp}{14bp}\selectfont #1}
\newcommand{\TickText}[1]{\fontsize{10.2bp}{11.4bp}\selectfont #1}
\newcommand{\SmallText}[1]{\fontsize{9.2bp}{10.4bp}\selectfont #1}
\newcommand{\TagText}[1]{\fontsize{8.3bp}{9.4bp}\selectfont #1}
\newcommand{\TagBold}[1]{\fontsize{8.6bp}{9.7bp}\selectfont\bfseries #1}
\newcommand{\SBSetCoord}[2]{%
  \pgfmathsetmacro{\SBpx}{\PlotL + (#1-1)/(\MaxRank-1)*(\PlotR-\PlotL)}%
  \pgfmathsetmacro{\SBpy}{\PlotB - (#2-1)/(\MaxRank-1)*(\PlotB-\PlotT)}%
}
\newcommand{\SBOtherPoint}[3]{%
  \SBSetCoord{#2}{#3}%
  \coordinate (p-#1) at (\SBpx,\SBpy);%
  \filldraw[fill=pointfill,fill opacity=.67,draw=pointedge,draw opacity=.85,line width=.72bp]%
    (\SBpx,\SBpy) circle[radius=3.65bp];%
}
\newcommand{\SBCorePoint}[3]{%
  \SBSetCoord{#2}{#3}%
  \coordinate (p-#1) at (\SBpx,\SBpy);%
  \filldraw[fill=corefill,draw=coreedge,line width=1.05bp]%
    (\SBpx,\SBpy) circle[radius=4.25bp];%
}
\newcommand{\SBOursPoint}[3]{%
  \SBSetCoord{#2}{#3}%
  \coordinate (p-#1) at (\SBpx,\SBpy);%
  \fill[frontier,opacity=.20] (\SBpx,\SBpy) circle[radius=18bp];%
  \draw[frontier,line width=2.2bp,opacity=.85] (\SBpx,\SBpy) circle[radius=12.5bp];%
  \path[fill=oursfill,draw=oursedge,line width=2.05bp,line join=round]%
    (\SBpx,\SBpy-10.8)--(\SBpx-2.43,\SBpy-3.34)--
    (\SBpx-10.27,\SBpy-3.34)--(\SBpx-3.92,\SBpy+1.27)--
    (\SBpx-6.35,\SBpy+8.74)--(\SBpx,\SBpy+4.12)--
    (\SBpx+6.35,\SBpy+8.74)--(\SBpx+3.92,\SBpy+1.27)--
    (\SBpx+10.27,\SBpy-3.34)--(\SBpx+2.43,\SBpy-3.34)--cycle;%
}
\tikzset{%
  tag/.style={fill=white,fill opacity=.94,text opacity=1,draw=tagborder,line width=.65bp,%
    rounded corners=2.2bp,inner xsep=4.2bp,inner ysep=2.5bp,text=spine},%
  ourtag/.style={fill=oursedge,draw=oursedge,line width=.65bp,rounded corners=2.2bp,%
    inner xsep=4.5bp,inner ysep=2.7bp,text=white},%
  leader/.style={draw=secondary,line width=.65bp,opacity=.78,line cap=round}%
}
\begin{tikzpicture}[x=1bp,y=-1bp,every node/.style={inner sep=0bp,outer sep=0bp}]
\path[use as bounding box] (0,0) rectangle (\PageW,\PageH);
\clip (0,0) rectangle (\PageW,\PageH);
\fill[white] (0,0) rectangle (\PageW,\PageH);
\fill[panelbg] (\PlotL,\PlotT) rectangle (\PlotR,\PlotB);
\foreach \rr in {10,20,30,40,50} {%
  \SBSetCoord{\rr}{\rr}%
  \draw[gridc,line width=.72bp] (\SBpx,\PlotT)--(\SBpx,\PlotB);%
  \draw[gridc,line width=.72bp] (\PlotL,\SBpy)--(\PlotR,\SBpy);%
  \node[anchor=north,text=secondary] at (\SBpx,\PlotB+7) {\TickText{\rr}};%
  \node[anchor=east,text=secondary] at (\PlotL-8,\SBpy) {\TickText{\rr}};%
}
\SBSetCoord{1}{1}
\node[anchor=north,text=secondary] at (\SBpx,\PlotB+7) {\TickText{1}};
\node[anchor=east,text=secondary] at (\PlotL-8,\SBpy) {\TickText{1}};
\draw[spine,line width=.95bp] (\PlotL,\PlotT)--(\PlotL,\PlotB)--(\PlotR,\PlotB);
\draw[diagc,line width=1.05bp,dash pattern=on 5bp off 3.2bp]%
  (\PlotL,\PlotB)--(\PlotR,\PlotT);
\node[rotate=-27,anchor=south,text=diagc] at (392,194)%
  {\SmallText{rank invariance}};
\node[anchor=base,text=spine] at ({(\PlotL+\PlotR)/2},414)%
  {\AxisText{$R^2$ mean rank (lower is better)}};
\node[rotate=90,anchor=base,text=spine] at (20,{(\PlotT+\PlotB)/2})%
  {\AxisText{CRPS mean rank (lower is better)}};
\node[anchor=north west,text=secondary] at (\PlotL+10,\PlotT+9)%
  {\SmallText{$\swarrow$ better on both metrics}};
\SBOtherPoint{tabpfn_v3}{10.029412}{6.156863}%
\SBOtherPoint{tabpfn_realv2_5}{20.696078}{19.960784}%
\SBOtherPoint{crepes_tabiclv2}{23.264706}{29.441176}%
\SBOtherPoint{crepes_tabiclv2_mondrian}{29.441176}{31.460784}%
\SBOtherPoint{catboost_quantile}{33.519608}{34.617647}%
\SBOtherPoint{tabm_d}{35.950980}{33.794118}%
\SBOtherPoint{pytabkit_realmlp_td}{36.137255}{34.049020}%
\SBOtherPoint{ngboost_gaussian}{38.058824}{38.950980}%
\SBOtherPoint{xgb_vector}{38.833333}{38.764706}%
\SBOtherPoint{crepes_catboost_difficulty}{39.578431}{40.107843}%
\SBOtherPoint{xgblss_Gaussian}{39.833333}{41.411765}%
\SBOtherPoint{cde_MDN}{41.921569}{41.009804}%
\SBOtherPoint{cde_NF}{42.323529}{41.617647}%
\SBOtherPoint{crepes_catboost_difficulty_mondrian}{42.607843}{42.882353}%
\SBOtherPoint{crepes_xgb_difficulty}{43.019608}{42.186275}%
\SBOtherPoint{nflows_rqs}{44.137255}{44.794118}%
\SBOtherPoint{crepes_xgb_difficulty_mondrian}{45.029412}{44.901961}%
\SBOtherPoint{surjectors_maf}{45.656863}{46.029412}%
\SBCorePoint{nori_30m}{14.901961}{11.588235}%
\SBCorePoint{tabiclv2}{15.990196}{11.882353}%
\SBCorePoint{tabpfn_v2_6}{16.872549}{18.892157}%
\SBCorePoint{nori}{20.098039}{17.764706}%
\SBCorePoint{forest_diffusion_flow}{35.382353}{39.529412}%
\SBCorePoint{xgb_vector_quantile}{40.196078}{41.764706}%
\SBCorePoint{flexcode_randomforest}{42.401961}{38.803922}%
\SBCorePoint{pymc_bart}{43.941176}{46.823529}%
\SBCorePoint{flexcode_xgboost}{45.127451}{41.588235}%
\SBOursPoint{exaonetabular}{8.754902}{6.068627}%
\node[ourtag,anchor=west] (tag-exaonetabular) at (85,307)
  {\TagBold{exaonetabular}};%
\draw[leader] (p-exaonetabular)--(tag-exaonetabular);%
\node[tag,anchor=east] (tag-nori_30m) at (193,272)
  {\TagText{nori\_30m}};%
\draw[leader] (p-nori_30m)--(tag-nori_30m);%
\node[tag,anchor=west] (tag-tabiclv2) at (302,302)
  {\TagText{tabiclv2}};%
\draw[leader] (p-tabiclv2)--(tag-tabiclv2);%
\node[tag,anchor=east] (tag-tabpfn_v2_6) at (212,226)
  {\TagText{tabpfn\_v2\_6}};%
\draw[leader] (p-tabpfn_v2_6)--(tag-tabpfn_v2_6);%
\node[tag,anchor=west] (tag-nori) at (337,261)
  {\TagText{nori}};%
\draw[leader] (p-nori)--(tag-nori);%
\node[tag,anchor=east] (tag-forest_diffusion_flow) at (414,88)
  {\TagText{forest\_diffusion\_flow}};%
\draw[leader] (p-forest_diffusion_flow)--(tag-forest_diffusion_flow);%
\node[tag,anchor=east] (tag-flexcode_randomforest) at (492,136)
  {\TagText{flexcode\_randomforest}};%
\draw[leader] (p-flexcode_randomforest)--(tag-flexcode_randomforest);%
\node[tag,anchor=east] (tag-xgb_vector_quantile) at (468,61)
  {\TagText{xgb\_vector\_quantile}};%
\draw[leader] (p-xgb_vector_quantile)--(tag-xgb_vector_quantile);%
\node[tag,anchor=east] (tag-flexcode_xgboost) at (625,146)
  {\TagText{flexcode\_xgboost}};%
\draw[leader] (p-flexcode_xgboost)--(tag-flexcode_xgboost);%
\node[tag,anchor=east] (tag-pymc_bart) at (635,57)
  {\TagText{pymc\_bart}};%
\draw[leader] (p-pymc_bart)--(tag-pymc_bart);%
\filldraw[fill=pointfill,fill opacity=.67,draw=pointedge,line width=.72bp]%
  (110,23) circle[radius=3.7bp];
\node[anchor=base west,text=spine] at (120,27) {\SmallText{retained model}};
\filldraw[fill=corefill,draw=coreedge,line width=1bp]%
  (239,23) circle[radius=4.2bp];
\node[anchor=base west,text=spine] at (250,27) {\SmallText{labelled model}};
\draw[frontier,line width=2bp] (383,23) circle[radius=7bp];
\node[anchor=base west,text=spine] at (396,27) {\SmallText{Pareto-optimal}};
\draw[diagc,line width=1bp,dash pattern=on 4bp off 2.8bp] (517,23)--(543,23);
\node[anchor=base west,text=spine] at (551,27) {\SmallText{rank invariance}};
\node[anchor=base east,text=secondary] at (678,44)%
  {\SmallText{28 models / 102 datasets}};
\end{tikzpicture}%
\endgroup%
%
    }
    \caption{
        \textbf{Joint point-estimation and predictive-distribution rankings on ScoringBench.}
        Each point represents a model with valid results for both metrics and is positioned by its official Autorank mean ranks over 102 regression datasets; lower ranks are better on both axes. Fine-tuned and HPO variants are omitted for readability, leaving 28 models. The dashed diagonal indicates identical rank under $R^2$ and CRPS, while the highlighted Pareto-optimal point identifies \EXAONETabular{}, which achieves the best mean rank for both metrics. Name tags are restricted to selected representative models.}
    \label{fig:scoringbench-results}
\end{figure}
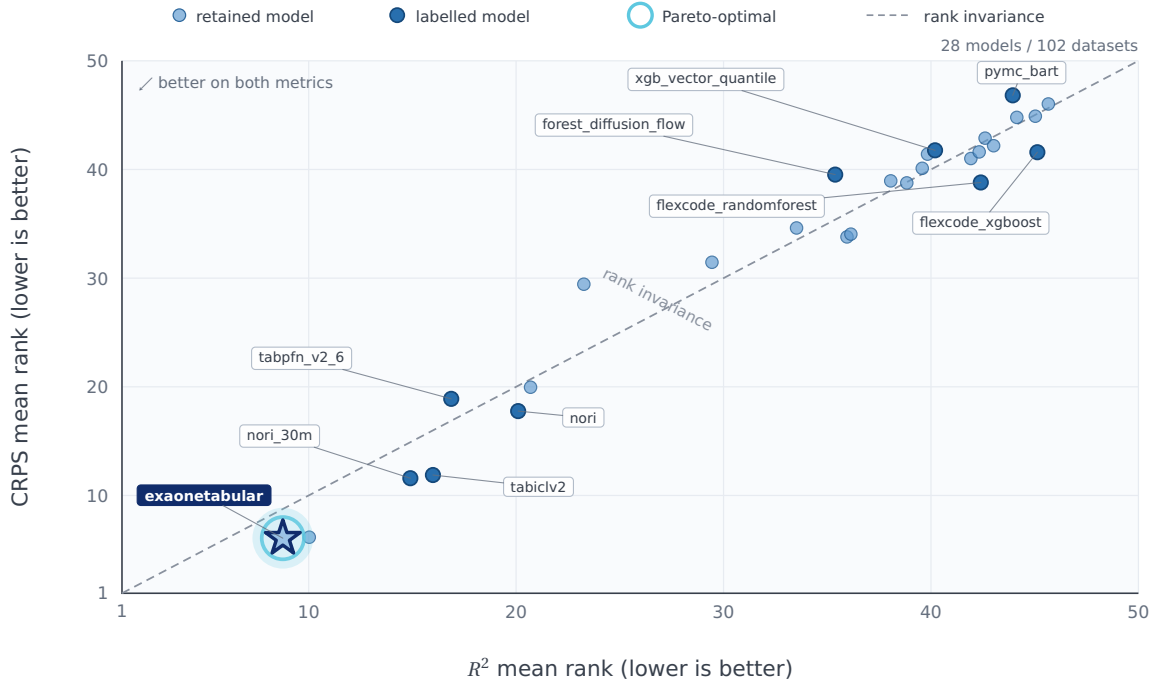

ScoringBench complements TabArena, BCCO, and TALENT by evaluating an aspect of regression performance that is not captured by point estimates alone. Whereas $R^2$ and RMSE assess the quality of a scalar prediction derived from the predictive distribution, CRPS evaluates the complete predictive cumulative distribution. The benchmark therefore tests whether a model that provides accurate point predictions also represents the uncertainty and dispersion of possible target values appropriately.

\paragraph{Point Estimation}
Figure~\ref{fig:scoringbench-results} jointly compares the official leaderboard mean ranks for $R^2$ and CRPS. Along the $R^2$ axis, \EXAONETabular{} achieves the best mean rank, indicating the strongest aggregate point-estimation performance across the evaluated regression datasets. It also ranks first under RMSE, showing that the result is consistent across the two squared-error-based point-prediction metrics rather than being specific to the normalization used by $R^2$.

\paragraph{Predictive Distribution}
The same figure shows that \EXAONETabular{} also achieves the best mean rank under CRPS. Unlike $R^2$ and RMSE, which evaluate the predictive mean, CRPS assesses the full predictive distribution. The joint position of \EXAONETabular{} at the leading corner therefore indicates that its strong point-estimation performance is accompanied by leading distributional performance.

The leading CRPS rank therefore shows that the point-estimation performance of \EXAONETabular{} is not obtained at the expense of distributional quality. Its predictive distributions achieve the strongest aggregate agreement with the observed outcomes among the eligible models, while its predictive means simultaneously lead the $R^2$ and RMSE comparisons. Because CRPS combines multiple aspects of probabilistic prediction, this result should be interpreted as the best overall CRPS performance on the benchmark rather than as a standalone claim of perfect calibration. Together, these results demonstrate that \EXAONETabular{} provides both accurate point estimates and high-quality distributional predictions across heterogeneous tabular regression tasks.

\section{Limitations and Future Work}
\label{sec:limitations}

\paragraph{Class-Count Handling.}

The native classification head supports up to 10 classes. Datasets with larger label spaces are handled through an ECOC-based decomposition at inference time. This procedure requires multiple classification subproblem predictions and therefore increases inference cost as the number of classes grows. A class-count-independent prediction head is a potential direction for future work.

\paragraph{Large-Context Inference.}
\label{sec:large-context}

Query chunking controls peak query-side memory because query predictions are conditionally independent given a fixed support context and compatible feature encoding. The inference pipeline can reuse support-side keys and values across query chunks when applicable, reducing redundant support-side computation. Otherwise, it uses uncached execution and recomputes the corresponding support-side values for each chunk. Although caching can improve inference efficiency, it introduces an additional memory cost, and its benefit depends on the support size, query size, and execution environment.

Support sets beyond the configured inference limit are currently subsampled. Potential future directions include context compression, representative-context selection, clustering-based support reduction, retrieval-based context construction, and adaptive support-set sampling. These methods require systematic evaluation of the trade-offs among inference latency, memory consumption, support compression, and predictive performance.






\section{License and Permitted Use}
\label{sec:license}

The \EXAONETabular{} release is distributed under separate licenses for the inference software and the model weights.

The \code{exaonetabular} inference runtime and associated source code released through the official GitHub repository are provided under the BSD-3-Clause-LG AI Research License. Subject to the terms of that license, the software may be used, modified, and redistributed, including for commercial purposes.

The released \EXAONETabular{} model weights are licensed separately under the \emph{EXAONE AI Model License Agreement 1.2 - NC}. The model license permits use of the weights for non-commercial research purposes, subject to the terms and restrictions specified in the license agreement. Commercial use of the released model weights requires separate authorization from the licensor.

The complete and authoritative license terms are distributed with the corresponding software and model weights through the official GitHub and Hugging Face repositories. This section provides only a high-level summary and does not replace, modify, or supersede those license terms.
\section{Conclusion}
\label{sec:conclusion}

We introduced \EXAONETabular{}, a compact tabular foundation model family for classification and regression based on in-context learning. Its central architectural contribution is the Cross-Axis Summary Transformer (CAST), which interleaves within-row feature processing with support-conditioned across-row contextualization throughout the Transformer hierarchy. Distinct item-summary and feature-summary tokens mediate these interactions: item summaries join feature-axis sequences, while feature summaries join support item-axis sequences, allowing the model to retain cell-level representations while progressively refining feature interactions with support-set context.

The classification and regression models are trained separately on synthetic tasks generated entirely from an SCM prior and perform prediction without dataset-specific gradient updates. The resulting models contain approximately 20.8M and 21.1M parameters, respectively, while supporting native missing-value handling, task-specific prediction heads, test-time ensembling, and support-conditioned chunked inference.

Across four public benchmarks, \EXAONETabular{} demonstrates consistently strong performance across classification, point regression, and probabilistic regression. On TabArena, the classification model ranks first among all evaluated methods using a single untuned default configuration, while regression reaches the performance regime of the 1.64B-parameter TabFM at roughly 1/11 the inference cost. On BCCO and TALENT, \EXAONETabular{} ranks second in classification and first in regression. On ScoringBench, it achieves the best mean rank for $R^2$, RMSE, and CRPS, demonstrating leading performance for both point estimation and predictive-distribution quality. Together, these results establish \EXAONETabular{} as a compact and highly competitive foundation model family across a broad range of real-world tabular prediction settings.

\appendix


\FloatBarrier
\printbibliography
\end{document}